%% file: neurips_data_2024.tex
\documentclass{article}

\PassOptionsToPackage{numbers, compress}{natbib}

\usepackage[preprint]{neurips_data_2024}

\usepackage[utf8]{inputenc} 
\usepackage[T1]{fontenc}    
\usepackage{hyperref}       
\usepackage{url}            
\usepackage{booktabs}       
\usepackage{amsfonts}       
\usepackage{nicefrac}       
\usepackage{microtype}      
\usepackage{xcolor}         

\definecolor{caoblue}{HTML}{242671}
\definecolor{dullred}{HTML}{AA0000}
\definecolor{taskcolor}{HTML}{B85450}
\definecolor{instcolor}{HTML}{9673A6}
\definecolor{actcolor}{HTML}{6C8EBF}
\definecolor{linegrey}{HTML}{999999}

\usepackage{amsmath}
\usepackage{amssymb}
\usepackage{pifont}
\usepackage{textcomp}

\usepackage{graphicx}
\usepackage[para]{threeparttable}
\usepackage{array}
\usepackage{wrapfig}
\usepackage{multirow}

\usepackage[inline]{enumitem}
\usepackage{longtable}

\usepackage{subfigure}
\usepackage{caption}
\usepackage{float}
\usepackage{newfloat}
\usepackage{listings}
\DeclareCaptionStyle{ruled}{labelfont=normalfont,labelsep=colon,strut=off} 
\floatstyle{ruled}
\newfloat{listing}{tb}{lst}{}
\floatname{listing}{Listing}

\usepackage{cleveref}

\crefname{section}{\S}{\S\S}
\Crefname{section}{\S}{\S\S}
\crefname{appendix}{\S}{\S\S}
\Crefname{appendix}{\S}{\S\S}
\crefname{figure}{Fig.}{Figs.}
\Crefname{figure}{Fig.}{Figs.}
\crefname{table}{Tab.}{Tabs.}
\Crefname{table}{Tab.}{Tabs.}

\newcommand{\XXX}{Qualified}
\newcommand{\xxx}{qualified}

\title{Mobile-Env: Building \XXX{} Evaluation Benchmarks for LLM-GUI Interaction}

\author{%
	Danyang Zhang$^1$ \quad
    Zhennan Shen$^1$ \quad
    Rui Xie$^1$ \quad
    Situo Zhang$^1$ \quad
    Tianbao Xie$^2$ \\
    \textbf{Zihan Zhao}$^1$ \quad
    \textbf{Siyuan Chen}$^1$ \quad
 	\textbf{Lu Chen}$^1$ \quad
	\textbf{Hongshen Xu}$^1$ \quad
	\textbf{Ruisheng Cao}$^1$ \quad
	\textbf{Kai Yu}$^1$ \\
    $^1$Shanghai Jiao Tong University, Shanghai, China \\
    $^2$The University of Hong Kong, Hong Kong, China \\
	\texttt{zhang-dy20@sjtu.edu.cn} \\
}

\begin{document}

\maketitle

\begin{abstract}
The Graphical User Interface (GUI) is pivotal for human interaction with the
digital world, enabling efficient device control and the completion of complex
tasks.  Recent progress in Large Language Models (LLMs) and Vision Language
Models (VLMs) offers the chance to create advanced GUI agents. To ensure their
effectiveness, there's a pressing need for \xxx{} benchmarks that provide
\textit{trustworthy} and \textit{reproducible} evaluations --- a challenge
current benchmarks often fail to address.  To tackle this issue,  we introduce
\textbf{Mobile-Env}, a comprehensive toolkit tailored for creating GUI
benchmarks in the Android mobile environment.  Mobile-Env offers an
\emph{isolated and controllable setting} for \emph{reliable evaluations}, and
accommodates \emph{intermediate instructions and rewards} to reflect real-world
usage more naturally.  Utilizing Mobile-Env, we collect an open-world task set 
across various real-world apps and a fixed world set, WikiHow, which captures a
significant amount of dynamic online contents for fully controllable and
reproducible evaluation.  We conduct comprehensive evaluations of LLM agents
using these benchmarks. Our findings reveal that even advanced models (e.g.,
GPT-4V and LLaMA-3) struggle with tasks that are relatively simple for humans.
This highlights a crucial gap in current models and underscores the importance
of developing more capable foundation models and more effective GUI agent
frameworks.\footnote{Mobile-Env is open-sourced at
\url{https://github.com/X-LANCE/Mobile-Env}. WikiHow task set is publicly
available at \url{https://huggingface.co/datasets/X-LANCE/WikiHow-taskset}.}
\end{abstract}


\section{Introduction}


\begin{figure}
    \centering
    \includegraphics[width=0.9\linewidth]{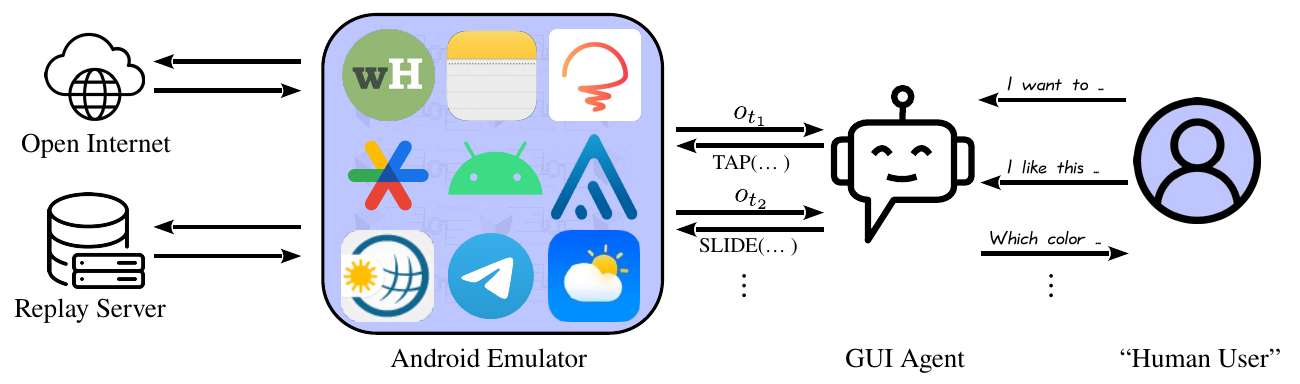}
    \caption{Mobile-Env platform. Mobile-Env comprises an Android emulator and a simple ``real user'' simulator
    (the purple sections).
    The Android emulator can host various real-world apps and run them in either open Internet or strictly 
    controlled environments, \textit{i.e.}, the replay server of crawled resources. The agent interacts with the
    emulator conditioned by the user's instructions. Intermediate instructions may be received during execution
    to supplement task information or user intents.}
    \label{fig:mob_env_feat}
\end{figure}

Graphical User Interface (GUI) serves as the main approach for humans to interact
with digital devices like mobile phones, personal computers, and web pages.
People perform a vast number of tasks through GUIs, such as web browsing, article reading, ticket booking,
and file revision. Most of these tasks are time-consuming and repetitive. By employing powerful autonomous 
agents for these GUI tasks, we can significantly alleviate workloads and boost productivity. 
Recent Large Language Models (LLMs) and Vision Language Models (VLMs)~\citep{LongOuyang2022NeurIPS_InstructGPT,OpenAI2023_GPT4,OpenAI2023_GPT4V,AIMeta2024_LLaMA3,ZhengxiaoDu2022ACL_GLM,WeihanWang2023_CogVLM} demonstrate impressive capabilities in understanding both natural and structured texts as well 
as multimodal contents. They show great potential for interacting with complex multimodal structural GUI environments. 
The advancement of these foundation models presents a promising opportunity to develop general-purpose GUI agents.
Consequently, to fairly assess the capabilities of a wide array of LLM\slash VLM-based agents, 
it is essential to build a benchmark for GUI interaction that can thoroughly and reliably evaluate agents 
in a controllable manner.

Currently, a large portion of benchmarks for GUI interaction are static 
datasets~\citep{LiangtaiSun2022EMNLP_META_GUI,XiangDeng2023_Mind2Web,RunliangNiu2024_ScreenAgent,XingHanLu2024_WebLINX}. 
These datasets provide fixed ground-truth annotations and evaluate the model's performance through action sequence matching. Therefore, trajectories with alternative routes or detours while still completing the task would be mistakenly marked as failures, leading to wrong evaluations.
Interactive benchmarks like AndroidArena~\citep{MingzheXing2024_AndroidArena} adopt
similarity of action sequences just like static datasets, and thus fail to assess
agent performances adequately. Recent works on LLM-based GUI 
agents~\citep{ChiZhang2023_AppAgent,JunyangWang2024_MobileAgent,ChaoyunZhang2024_UFO}
conduct evaluations with the help of LLMs or human judgments. However, their experiments are conducted in completely open environments without any isolation or
controls, making the results unreproducible and incomparable.  

Furthermore, existing interactive environments lack two crucial features: intermediate rewards and intermediate instructions. Interactive environments should support more agent designing methods, such as Reinforcement Learning (RL)~\citep{PeterCHumphreys2022ICML_CCNet}. In these cases, intermediate rewards are essential for agents to learn efficiently in GUI environments with complex observations and extensive action space. Additionally, it is unrealistic for users to provide a comprehensive command containing all task details at once. Therefore, intermediate instructions should be provided throughout the task based on information from newly encountered screens, creating more natural and practical use cases.

In general, a \xxx{} evaluation benchmark for GUI interaction is expected to have several
features: 1) \textbf{reliable evaluation} that assessing the execution result rather than
the execution trajectory as multiple feasible trajectories may exist, 2) \textbf{isolated and controllable 
environments} to prevent agents' latent dangerous actions from corrupting real-world environments and
guarantee reproducible evaluation and fair comparison, 3) 
\textbf{intermediate rewards} for better supporting interactive learning methods, and 4)
\textbf{intermediate instructions} to face more natural and realistic use cases. 


Towards building \xxx{} evaluation benchmarks for GUI interaction, we developed
the Mobile-Env platform as a comprehensive toolkit for GUI benchmark
construction. Noticed that mobiles hold the most market share among consumer
electronic
devices\footnote{\url{https://gs.statcounter.com/platform-market-share/desktop-mobile-tablet/worldwide/2023}},
we build Mobile-Env based on the Android\texttrademark{} mobile platform.
Mobile-Env offers an isolated interactive environment and designs a flexible
framework to reliably evaluate the widest range of tasks.
Additionally, intermediate rewards in Mobile-Env enable efficient reinforcement
learning, and the intermediate instructions facilitate the development of
conversational agents.
With the help of Mobile-Env, \xxx{} evaluation benchmarks can be built much
more easily.

Based on Mobile-Env, we design 74 tasks across a number of real-world apps. We
further build a task set, WikiHow, containing 150 tasks generated from 13 templates, by fixing a remarkable
scale of dynamic online contents, achieving
absolutely controllable and reproducible evaluation. Extensive experiments are conducted 
on these designed tasks with current advanced LLMs and VLMs. It is found that even the most advanced models 
struggle with tasks that are relatively simple for humans. These results reveal the deficiency of current LLMs and 
VLMs as reliable and effective GUI assistants, underscoring the necessity for developing more capable foundation
models and knowledge-augmented agents.

\section{Background}

\subsection{Problem formulation}

As illustrated in \cref{fig:mob_env_feat}, a GUI agent $A$ receives the human user's instructions and 
completes the task in a GUI system (\textit{i.e.}, the Android emulator in the figure). The ``human user''
and the GUI system are regarded as the environment holding hidden intents\slash internal states $s_t$, which the agent
should deduce via observations. The observation $o_t$ the agent receives at each step consists of the screen information
from the GUI system
and an optional intermediate instruction from the ``human user''. Then it predicts an action $a_t$ to
operate the GUI system or to respond to the ``human user'', causing a state transition $(s_t, a_t, s_{t+1})$.
An optional reward can be given to assess the transition and can be used to improve the agent.
Observation receiving and action taking iterate alternatively in this manner until the task comes to success or failure.


\subsection{\XXX{} benchmarks}


\begin{table}[t]
    \centering
    \footnotesize
    \begin{threeparttable}
        \addtolength{\tabcolsep}{-3pt}
        \centering
        \caption{Comparison of Mobile-Env with existing GUI benchmarks. The table shows the support of the 
        benchmarks\slash benchmark toolkits for result-based evaluation, truly
        controllable environments, intermediate rewards, and intermediate instructions.}
        \label{tab:plf_cmp}
        \begin{tabular}{ccm{1.8cm}<{\centering}m{2cm}<{\centering}m{1.8cm}<{\centering}m{1.8cm}<{\centering}}
            \toprule[1.5pt]
            & Benchmark (Toolkit) & Result-based Evaluation & Controllable Environments & Intermediate Rewards & Intermediate Instructions \\
            \midrule
            \multirow{5}{*}{\rotatebox[origin=c]{90}{\parbox[c]{1.5cm}{\centering\noindent Static Dataset}}} & PixelHelp~\citep{YangLi2020ACL_PixelHelp} & \ding{55} & - & - & \ding{55} \\
            & META-GUI~\citep{LiangtaiSun2022EMNLP_META_GUI} & \ding{55} & - & - & \ding{51} \\
            & Mind2Web~\citep{XiangDeng2023_Mind2Web} & \ding{55} & - & - & \ding{55} \\
            & AitW~\citep{ChristopherRawles2023_AitW} & \ding{55} & - & - & \ding{55} \\
            & WebLINX~\citep{XingHanLu2024_WebLINX} & \ding{55} & - & - & \ding{51} \\
            \midrule
            \multirow{9}{*}{\rotatebox[origin=c]{90}{\parbox[c]{2cm}{\centering\noindent Interactive Environment}}} & MiniWoB++~\citep{EvanZheranLiu2018ICLR_MiniWoBpp} & \ding{51} & \ding{51} & \ding{55} & \ding{55} \\
            & AndroidEnv~\citep{DanielToyama2021_AndroidEnv} & \ding{51} & \ding{55} & \ding{51} & \ding{55} \\
            & WebShop~\citep{ShunyuYao2022_WebShop} & \ding{51} & \ding{51} & \ding{55} & \ding{55} \\
            & WebArena~\citep{ShuyanZhou2023_WebArena} & \ding{51} & \ding{51} & \ding{55} & \ding{55} \\
            & VisualWebArena~\citep{JingYuKoh2024_VisualWebArena} & \ding{51} & \ding{51} & \ding{55} & \ding{55} \\
            & AndroidArena~\citep{MingzheXing2024_AndroidArena} & \ding{55} & \ding{55} & \ding{55} & \ding{55} \\
            & OSWorld~\citep{TianbaoXie2024_OSWorld} & \ding{51} & \ding{55} & \ding{55} & \ding{55} \\
            & AndroidWorld~\citep{RawlesChristopher2024_AndroidWorld} & \ding{51} & \ding{55} & \ding{55} & \ding{55} \\
            \cmidrule{2-6}
            & Mobile-Env (Ours) & \ding{51} & \ding{51} & \ding{51} & \ding{51} \\
            \bottomrule[1.5pt]
        \end{tabular}
    \end{threeparttable}
\end{table}

This subsection will discuss the status quo of existing GUI benchmarks and the proposed features of a \xxx{}
benchmark. An overall comparison is available in \cref{tab:plf_cmp}.


\begin{figure}
    \centering
    \includegraphics[width=0.95\linewidth]{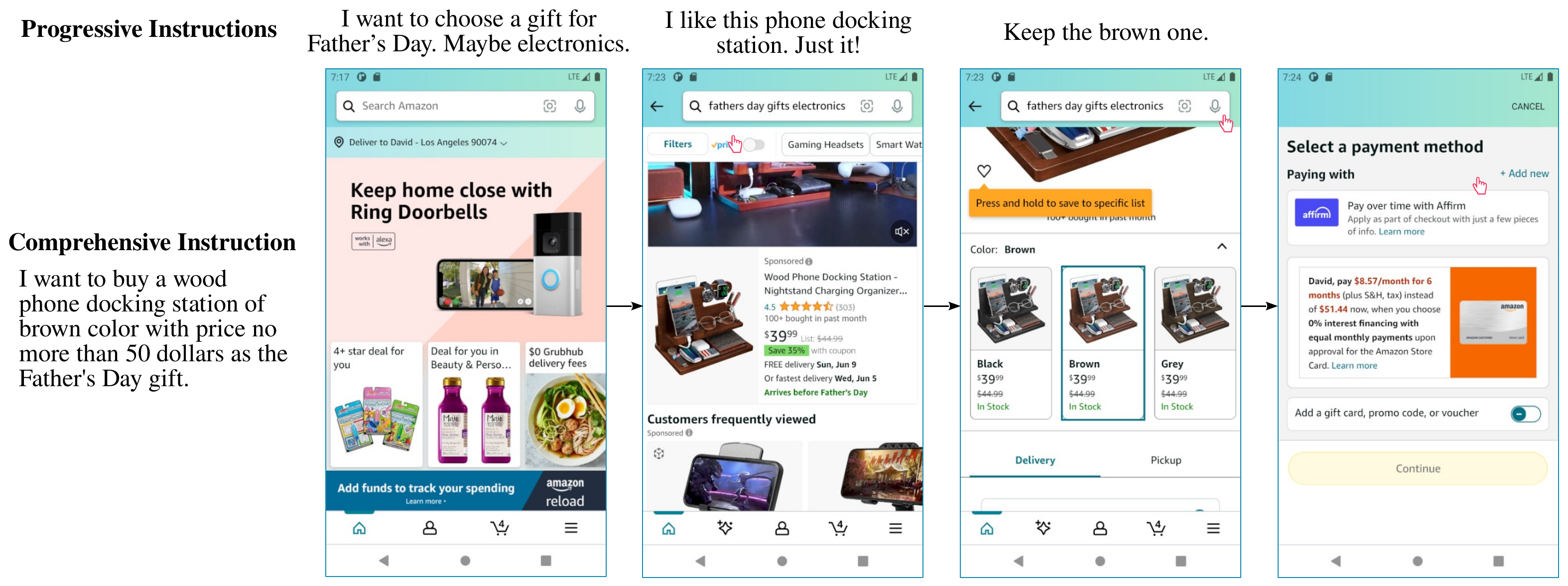}
    \caption{Progressive instructions vs.\ comprehensive instruction. A single comprehensive instruction that covers every detail is usually verbose and unnatural. In real use cases, the user's intent may not be fully clear at the beginning but becomes more defined during the execution process.}
    \label{fig:inter_instruction}
\end{figure}

\paragraph{Reliable evaluation} Existing GUI interaction benchmarks put
a lot of efforts into evaluating 
the agent's executions. Most static 
benchmarks~\citep{YangLi2020ACL_PixelHelp,LiangtaiSun2022EMNLP_META_GUI,XiangDeng2023_Mind2Web,ChristopherRawles2023_AitW,RunliangNiu2024_ScreenAgent,XingHanLu2024_WebLINX} 
simply perform step-wise evaluation by comparing the predicted actions
with the reference action trajectory.
However, there are usually multiple feasible solutions for a specific task goal. 
The agent may also conduct explorations not directly leading to task completion. In order to conduct reliable evaluations, the benchmark
should inspect the execution \emph{result} rather than the execution \emph{process}
\citep{ShuyanZhou2023_WebArena,JingYuKoh2024_VisualWebArena}.
Specifically, the task goal can always be viewed as a subset of hidden states.
As long as the agent reaches these states, its execution will be deemed successful.
The hidden final state is supposed to be estimated through a series of system signals. 
Usually, a single type of signal cannot cover all the state transitions, \textit{e.g.},
several changes like form submission can only be reflected on GUI, but
not in the system logs adopted in \citet{DanielToyama2021_AndroidEnv} or files and
databases adopted in 
\citet{RawlesChristopher2024_AndroidWorld}. 
Therefore, diverse system signals should be leveraged to comprehensively assess
the final states \citep{TianbaoXie2024_OSWorld}. Mobile-Env exploits a wide range of
signal types,
including screen text, View Hierarchy (VH)\footnote{A tree-shape 
representation of the Android screens, similar to HTML of the web pages.}, 
system logs, and agent's Response to Human User (RHU). Furthermore, Mobile-Env designs flexible mechanisms
to combine multiple signals, achieving reliable and flexible evaluation.

\paragraph{Isolated \& controllable environments} Due to the lack of \xxx{} 
benchmarks, recent work on LLM-based GUI 
agents~\citep{DifeiGao2023_AssistGUI,ChiZhang2023_AppAgent,JunyangWang2024_MobileAgent,ChaoyunZhang2024_UFO}
conducts evaluations with real devices in fully open environments without 
any isolation or control. This approach leads to several issues: safety risks, unreproducible evaluations, and unfair comparisons.
\emph{Safety risks} are referred to as the fact that without environment \textbf{isolation},
the agent may mistakenly corrupt real-world devices or data resources. On the other
hand, many apps rely on online contents that vary at different
time and locations. Therefore, environments built on such apps will result in \emph{unreproducible evaluation}. 
Meanwhile, these factors will also affect the 
actual difficulty of the tasks in the environment
and result in \emph{unfair comparisons} between agents. 
For instance, an agent might easily find a Panda Express\footnote{A popular Chinese food restaurant in the USA.} in the USA but not in China.
Consequently, the only way to set up reproducible and
fair evaluation is to build a truly \textbf{controllable} environment by fixing the dynamic online contents 
and replaying during evaluations
\citep{TianlinShi2017ICML_MiniWoB,EvanZheranLiu2018ICLR_MiniWoBpp,ShuyanZhou2023_WebArena}. 
To enable common data replay for mobile apps, we reverse-engineered and developed a suite of solutions for replay server usage and app certificate unpinning.

\paragraph{Intermediate rewards}
Reinforcement Learning (RL) is a method of great potential for building intelligent 
agents. Extensive work is trying to leverage RL to build digital
agents~\citep{PeterCHumphreys2022ICML_CCNet,ShunyuYao2022NeurIPS_WebShop,YifeiZhou2024_ArCHer,ZhaiYuexiang2024_VLMRLAgent}. 
However, a common challenge in RL is reward sparsity, which is exacerbated by the increasingly complex observations and extensive action space in GUI environments. One way to alleviate this problem is to introduce intermediate
rewards~\citep{MaayanShvo2021CanadianAI_AppBuddy,JakeBruce2023ICLR_LearningProgressFromExpert,ChangMa2024_AgentBoard}.
Mobile-Env designs a listener-style evaluator and produces intermediate rewards safely for GUI interaction tasks.

\paragraph{Intermediate instructions} As shown in \cref{fig:inter_instruction}, a comprehensive task instruction covering all details is usually lengthy and verbose. It is impractical for real users to consider all variable factors at the outset. Realistic task instructions should be progressive. Initially, there is a simple instruction that outlines the user's basic intent. As the task progresses and new information emerges, the user's intent becomes clearer, and more details
are provided. Moreover, supporting intermediate instructions is crucial for evaluating conversational GUI agents~\citep{LiangtaiSun2022EMNLP_META_GUI,XingHanLu2024_WebLINX}, enabling agents and users to communicate about task details and progress.
To our knowledge, Mobile-Env is the first interactive environment that supports intermediate instructions, allowing for the evaluation of conversational GUI agents.

\section{Mobile-Env platform}

Mobile-Env offers a uniform framework to host various GUI environments (\cref{sub:host_gui}) and build \xxx{} GUI benchmarks (\cref{sub:build_bench}).
As it is a common way to ``work with the emulators
thus building these environments'', we will introduce the internal components and implementation details 
in the supplementary materials, and leave this section focusing on the use features of Mobile-Env.

\subsection{Universal hosting of GUI environments}
\label{sub:host_gui}


Mobile-Env is capable of hosting most real-world Android apps and 
offers uniform interfaces for all kinds of apps and mobile interaction tasks. To be specific, the observation of Mobile-Env
environments include both screenshots and VH (View Hierarchy) XML. The raw GUI action space of Mobile-Env
is pixel-wise atomic \verb|TOUCH| (finger down), \verb|LIFT| (finger up), and \verb|TEXT| (token input), ensuring a universal and fine-grained interface for interacting with various apps. 
Additionally, we provide a convenient way to use environment wrappers to modify the action space; for example, appending a \verb|LIFT| to a series of \verb|TOUCH| actions to create a \verb|TAP|. The RHU (Response to Human User) action is also available in Mobile-Env.
As shown in \cref{fig:mob_env_feat}, the agent can interact with Mobile-Env environments either in an open-world setting or within a strictly controlled world conditioned by multi-turn instructions.
This makes Mobile-Env a universal platform for various mobile interactive environments and agents.

\subsection{Creation of \xxx{} GUI benchmarks}
\label{sub:build_bench}

Mobile-Env is used to build \xxx{} GUI interaction benchmarks. 
Its first notable feature lies in the configuration setup: all the setup procedures, evaluation methods, and reward mechanisms are configured through an external task configuration file. This file adopts the text format of Protocol Buffer\footnote{https://protobuf.dev/reference/protobuf/textformat-spec/}, which is easier to comprehend and modify compared to complicated and overwrapped programming codes.
The subsequent part will introduce Mobile-Env's methodology towards building \xxx{} benchmarks by (1) implementing isolated and controllable environments, and (2) ensuring accurate task state estimation for reliable evaluation and immediate rewards \& instructions.

\begin{figure}[t]
    \centering
    \includegraphics[width=0.9\linewidth]{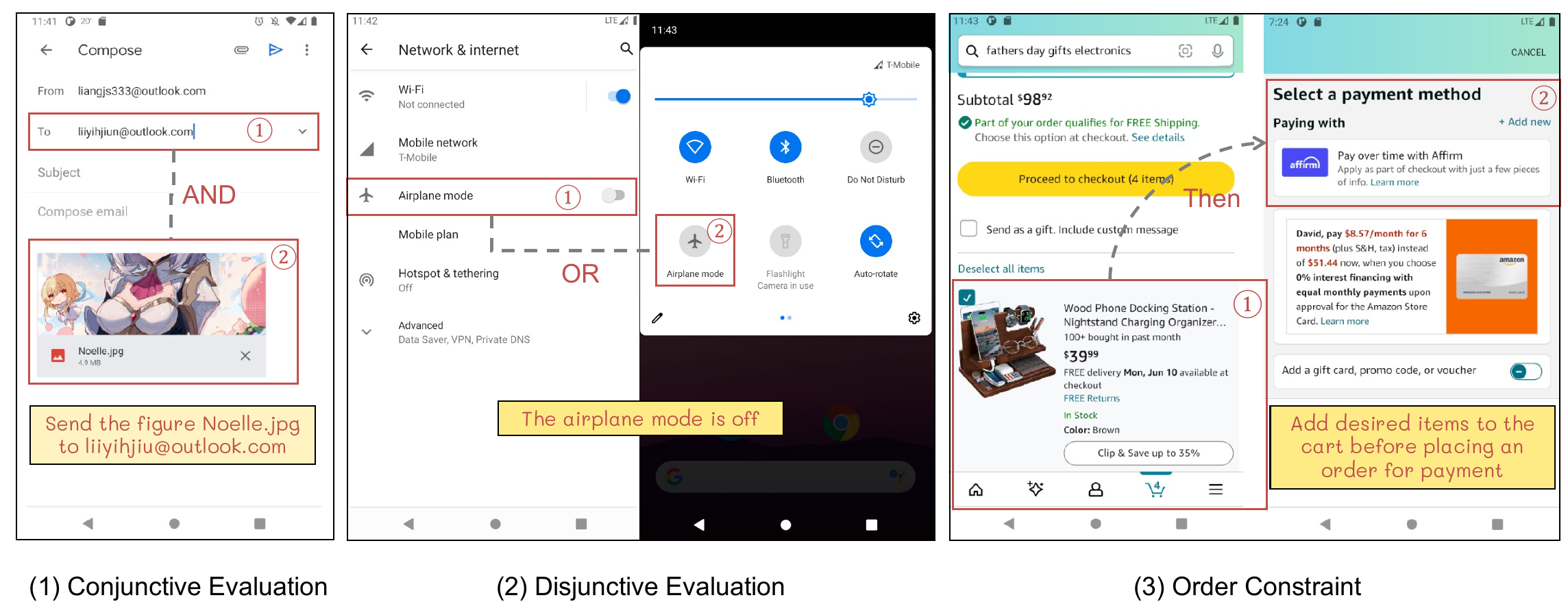}
    \caption{Mobile-Env utilizes three methods (\textit{i.e.}, conjunctive evaluation, disjunctive evaluation, and order constraints) for combining signals to achieve flexible evaluation.}
    \label{fig:sgn_combs}
\end{figure}



\subsubsection{Setup of isolated \& controllable environments} 
Mobile-Env sets up an environment using a predefined Android OS (Operating System) image and executes a sequence of ADB
(Android Debug Bridge) commands defined in the task configuration file. This process ensures that the environment is properly prepared for the task before awaiting the agent's execution.
If a fully controlled ``online'' environment needs to be established, a replay server for the crawled
app data should be launched in advance.
However, direct replay will face the challenge of certificate pinning in real-world apps, which prevents them from trusting SSL certificates issued by the replay proxy and leads to data replay failure.
To handle this problem, we implemented three solutions
according to the conclusions of reverse engineering, and verified these solutions on a series of common
real-world apps. Further details on these solutions are provided in the supplementary.

\subsubsection{Accurate task state estimation}
Accurate task state estimation is essential for both reliable evaluation and to guide the agent with intermediate rewards and instructions at the right time. 
Mobile-Env ensures this accuracy by utilizing a variety of signals including screen text, VH, system log, and RHU. Moreover, it introduces three strategies for combining these signals to enhance the flexibility of task state estimation (see \cref{fig:sgn_combs}): 
(1) \textbf{Conjunctive evaluation}: multiple signals must occur simultaneously, e.g., 
an e-mail is ready to be sent only if the recipient and contents are properly filled in.
(2) \textbf{Disjunctive evaluation}: multiple alternative signals are combined to avoid missing evidence of success. For instance, airplane mode on a mobile can be toggled from either the Quick Settings menu or the System Settings app. The final pages differ but both indicate whether airplane mode is on or off. (3) \textbf{Order constraint}: complex tasks consist of multiple stages, with some serving as prerequisites for others. For example, desired items must be added to the cart before placing an order for payment. 
By employing the aforementioned methods, Mobile-Env can achieve more accurate state estimation. Additional examples of task evaluation are detailed in the supplementary materials.

Furthermore, Mobile-Env adopts a ``listener-style'' approach for capturing task signals without disrupting the runtime state or the agent's actions. 
This design is different from the ``post-processing'' evaluators as described by \citet{TianbaoXie2024_OSWorld}, which might involve actions like saving an open file or closing an app that could potentially alter the environment state. The listener style adopted by Mobile-Env guarantees safe and correct state evaluations, as well as enables intermediate rewards and instructions, without interfering with the agent's ongoing execution.

\section{Task sets}
\label{sec:wikihow}

With the help of Mobile-Env, we collected two sets of GUI interaction tasks, one in
the open world, and another in a fixed controlled world. This section will briefly introduce
these two task sets.

\subsection{Open world set}

\begin{table}[t]
    \centering
    \footnotesize
    \begin{threeparttable}
        \centering
        \caption{Task instruction examples of the open world task set}
        \label{tab:op_task_eg}
        \begin{tabular}{p{8.5cm}ccc}
            \toprule[1.5pt]
            Instruction & Involved Apps & IR\tnote{\textasteriskcentered} & II\tnote{\dag} \\
            \midrule
            Turn on airplane mode for my phone. & System Settings & \ding{55} & \ding{55} \\
            Check my TOTP\tnote{\ddag} \ for GitHub and tell me the code. & Google Authenticator & \ding{55} & \ding{55}\\
            Help me to order a cup of my favorite milk tea of Coco (the store name). & Ele.me & \ding{51} & \ding{51} \\
            I have several funds and bills to be noted in OpenMoneyBox. Now open the app, enter the wizard, and help me to note the bills. & OpenMoneyBox & \ding{51} & \ding{51} \\
            Turn off my alarm in the morning if it will rain tomorrow, otherwise leave it on. & Weather, Alarm Clock & \ding{55} & \ding{55} \\
            \bottomrule[1.5pt]
        \end{tabular}
        \begin{tablenotes}
            \footnotesize
            \item[\textasteriskcentered] Intermediate Rewards
            \item[\dag] Intermediate Instructions
            \item[\ddag] Time-based One-Time Password
        \end{tablenotes}
    \end{threeparttable}
\end{table}

We define 74 GUI interaction tasks using real-world apps running directly in an open Internet environment to evaluate the agents' raw performance. Following AitW~\citep{ChristopherRawles2023_AitW} and AndroidWorld~\citep{RawlesChristopher2024_AndroidWorld}, we include app installation\slash uninstallation, clock settings, and Android system settings in the test set. Additionally, we select four popular apps and define a series of tasks based on their functions. The chosen apps are the instructional app WikiHow, the food delivery app Ele.me, and the review platforms Douban and IMDB.\footnote{WikiHow: \url{https://www.wikihow.com/Main-Page}. Ele.me: \url{https://www.ele.me/}. Douban: \url{https://www.douban.com/}. IMDB: \url{https://www.imdb.com/}.} We also define several tasks involving multiple apps to test the agents' ability to navigate across different applications. Examples of task instructions are shown in \cref{tab:op_task_eg}, and the full list of task instructions 
can be found in the supplementary materials.

\subsection{Fixed world set (WikiHow)}


\begin{wraptable}{R}{0.4\linewidth}
    \centering
    \caption{An example of sequential instructions in the fixed WikiHow task set}
    \label{tab:wikihow_eg}
    \begin{tabular}{p{5.25cm}}
        \toprule[1.5pt]
        1. How can I hide my ear gauges? \\
        2. {\it How to hide gauges} may be helpful. \\
        3. Share the article to other apps. \\
        \bottomrule[1.5pt]
    \end{tabular}
\end{wraptable}

Aiming at providing truly comparable results, we further collect a fixed world set based on the WikiHow app.
WikiHow is an app hosting
millions of ``how-to'' articles about step-by-step everyday tips. We crawled 107,448 pages containing 
856,045 text and multimedia resources from the WikiHow website to establish the fixed controlled environment.
Then, we craft
13 task configuration templates and generate 150 tasks for testing. The template-based instructions are further
rewritten with the help of ChatGPT\footnote{\url{https://chat.openai.com/}} to increase the diversity of
expression. Multiple rewritten candidates are recorded for each instruction and will be replaced randomly
at runtime.
An example of sequential instructions from the fixed WikiHow task set is shown in \cref{tab:wikihow_eg}. This task includes a series of sequential instructions guiding the agent to browse app contents. An intermediate reward is given upon completing each sub-instruction, and the episode is considered successful only when the final instruction is completed. Details of app data crawling, template crafting and instantiation, and instruction replacing are provided in the supplementary materials.


To draw deeper insights into the agents' capabilities, we split the fixed world
set into three categories: 1) 59 {\em
cross-page tasks} requiring pure navigation among pages, 2) 51 {\em in-page tasks}
engaging in several in-page operations, like bookmarking or voting
for an article, and 3) 40 {\em QA
tasks} requesting the agent to give a summarized answer according to the
article contents, \textit{e.g.}, ``read the article and summarize the needed
items for the task''. 

\section{Experiments}

\begin{wraptable}{R}{0.36\linewidth}
\centering
\footnotesize
\caption{Results on open world set}
\label{tab:op_wd_set_rlt}
\begin{tabular}{lcc}
    \toprule[1.5pt]
    \multicolumn{1}{c}{Model} & Rwd & SR \\
    \midrule
    Claude-3-Opus               & 0.86 & 32.43 \\
    Claude-3-Opus (V)           & 0.25 & 2.70  \\  
    GPT-4        & 1.09 & 43.24 \\
    GPT-4V      & 0.18 & 3.04  \\  
    \midrule
    AgentLM-70B               & 0.77 & 13.51 \\
    LLaMA-3-70B               & 1.11 & 40.54 \\
    \bottomrule[1.5pt]
\end{tabular}
\end{wraptable}

\subsection{Experiment settings}
\label{sub:exp_set}

We experiment with current leading LLMs and VLMs across three categories: (1) \textit{Closed-source LLMs} like GPT-3.5, GPT-4, and Claude-3, (2) \textit{Open-source LLMs} including LLaMA-2, LLaMA-3, and AgentLM of varying sizes, and (3) \textit{VLMs} such as GPT-4V and Claude-3-Opus.
Text-based LLMs take  a simplified
HTML derived from the VH observation as input. For VLMs, we provide the screenshot 
and Set-of-Marks (SoM)~\citep{JianweiYang2023_SoM}. 
Set-of-Marks are indexed visual tags annotating
the essential GUI elements on the image, helping the VLM to ground in the screenshot. We prompt the agents to first perform reasoning before the actual
action, following \citet{ShunyuYouICLR2023_ReAct}. The average reward (Rwd) through episode and success rate (SR, in \%)
are reported as the metrics.

\subsection{Results on open world set}
\label{sub:op_expe}

The results on the open world set are shown in \cref{tab:op_wd_set_rlt}, demonstrating the low performances across all the 
models, especially the VLMs (SRs of 2.70\% for Claude-3-Opus and 3.04\% for GPT-4V). This implies the deficiency
of current LLMs and VLMs as effective GUI assistants. Nevertheless, 
it is surprising to see a comparable performance of LLaMA-3-70B (SR of 40.54\%) with text GPT-4 (SR of 43.24\%), 
showing the incredible improvement of open-source models. 
Through carefully looking into the trajectories, we attribute the deficiency of the models, especially the VLMs, to 
the lack of the essential capabilities like common-sense GUI understanding and output format following.

\begin{wrapfigure}{R}{0.4\linewidth}
	\centering
	\includegraphics[width=\linewidth]{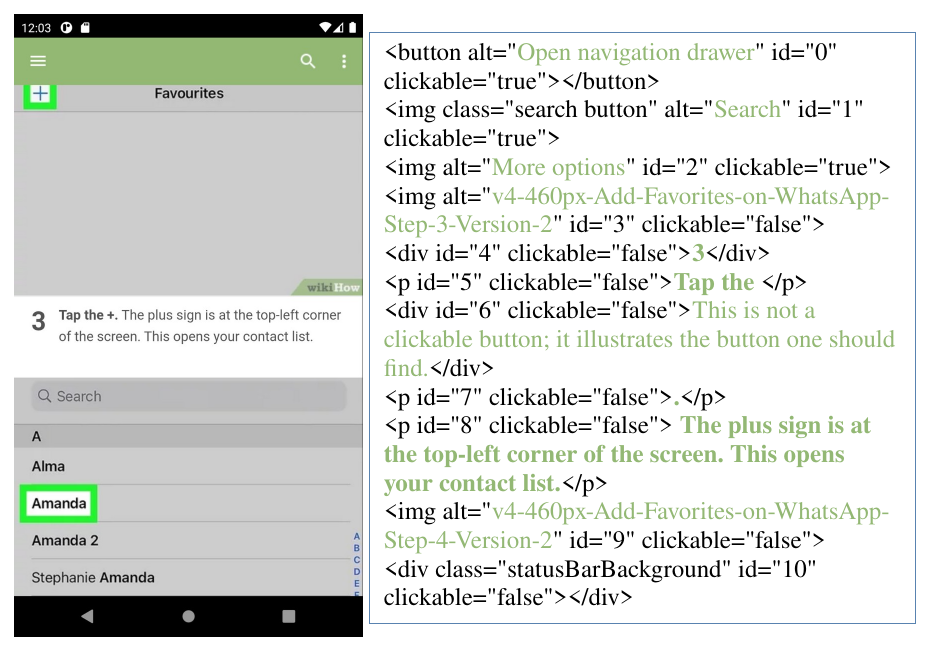}
	\caption{HTML input of QA tasks. The article cannot be perceived as whole
	and the text contents are interleaved with HTML markups.}
	\label{fig:qa_input}
\end{wrapfigure}

\paragraph{Common-sense knowledge of GUI understanding} 
GUI design follows user-friendly principles, employing familiar visual elements such as icons, buttons, and menus, which users can quickly recognize and interpret. In contrast, our experiments show that current LLMs and VLMs often lack this common sense and struggle to understand GUIs, especially on information-intensive screens like weather apps displaying
verbose forecasts for many future days.
For example, few agents can recognize the contents shown in File Manager and share a file from
it. And GPT-4 doesn't know how to display the receiving time of an e-mail. Specifically, we find that the VLMs 
perform much worse than text LLMs, as the screen texts and text prompts of icons are not as explicitly available
in the screenshot as in the HTML. Thus, the VLMs face more difficulties in understanding the texts, icons, and 
functions of the current screen, leading to terribly low performance.

\paragraph{Format following capabilities} Language model based agents are expected to output well-formatted texts so that
the actions can be easily parsed and executed. Our experiments show that even the most capable LLMs and VLMs
struggle to adhere to the output format strictly. Over half the predictions of GPT-4 and Claude-3-Opus contain
redundant ``new line'' characters. When input with screenshots, \textasciitilde 26.40\% outputs
of GPT-4V and \textasciitilde 10.87\% outputs
of Claude-3-Opus comprise only thoughts without actual actions. This problem has become a significant
bottleneck of agent's performance.

\begin{table}[t]
	\centering
    \footnotesize
	\caption{Experiment results of text LLMs on WikiHow task set}
	\label{tab:t_llm_rst}
	\begin{tabular}[b]{lcccccccc}
		\toprule[1.5pt]
		\multicolumn{1}{c}{\multirow{2}{*}{Model}} &	\multirow{2}{*}{Rwd} &	\multirow{2}{*}{SR} &	\multicolumn{2}{c}{Cross-Page} &	\multicolumn{2}{c}{In-Page} &	\multicolumn{2}{c}{QA} \\
		\cmidrule(lr){4-5}\cmidrule(lr){6-7}\cmidrule(lr){8-9}
		& & & Rwd &	SR &	Rwd &	SR &	Rwd &	SR \\
		\midrule
        Claude-3-Opus & 1.94 & 40.67 & 1.61 & 37.29 & 2.51 & 54.90 & 1.71 & 27.50 \\
		GPT-3.5 &	1.79 &	37.33 &	1.49 &	38.98 &	2.29 &	49.02 &	1.58 &	20.00 \\
        GPT-3.5 (1-turn) &	1.81 &	28.67 &	1.67 &	40.68 & 2.16 &	27.45 &	1.58 &	12.50 \\
		GPT-3.5-instruct &	1.76 &	36.67 &	1.71 &	44.07 &	2.02 &	35.29 &	1.50 &	27.50 \\
		GPT-4 &	2.01 &	43.33 &	1.95 &	61.02 &	2.35 &	50.98 &	1.65 &	7.50 \\
		\midrule
		LLaMA-2-7B &	0.65 &	4.00 &	0.81 &	10.17 &	0.78 &	0.00 &	0.25 &	0.00 \\
		LLaMA-2-13B &	1.08 &	11.33 &	1.14 &	27.12 &	1.37 &	0.00 &	0.63 &	2.50 \\
		LLaMA-2-70B &	1.49 &	20.00 &	1.44 &	30.51 &	1.65 &	7.84 &	1.34 &	20.00 \\
        \midrule
        LLaMA-3-8B & 1.74 & 30.67 & 1.58 & 35.59 & 1.90 & 23.53 & 1.77 & 32.50 \\
        LLaMA-3-70B & 2.10 & 47.33 & 1.83 & 49.15 & 2.47 & 54.90 & 2.02 & 35.00 \\
		\bottomrule[1.5pt]
	\end{tabular}
\end{table}


\subsection{Results of text LLMs on fixed world set}

As a supplement to results on the open world set, we further conduct experiments on 
the fixed world set to provide truly reproducible results.
The results of text LLMs on the WikiHow task set are presented in
\cref{tab:t_llm_rst}. It is observed that even the best performance
achieves a success rate of only 47.33\%, which highlights the deficiency of current LLMs
on GUI interactions again.

\paragraph{Performances across different task types} Overall, the QA tasks seem
more difficult than the other types of tasks.  This is not surprising, as the
QA tasks require the agent to scroll up and down to read the whole article and
extract the useful text contents out of the HTML markups. An example of the
HTML representation on the article page is shown in \cref{fig:qa_input}.  Thus,
this task is harder than simply seeking a specific button or hyperlink.  
Meanwhile, it is noticed that many models perform worse on the in-page tasks
than on the cross-page tasks. By inspecting the interaction trajectories, this
is attributed to that several in-page tasks necessitate actions like long-range
scrolling or checking into dropdown menus.  Successfully executing these
actions typically requires a strong belief in their effectiveness.  However,
many agents, lacking such confidence, usually try to click something else
during scrolling, aiming at reaching the target more quickly.  
Besides, there may be shortcuts that could simplify task completion, such as
clicking a reference list link to quickly navigate to the bottom of a page
instead of long-term scrolling.
However, few models possess such \textit{a priori} knowledge and manage to
complete the tasks in this manner.  How to establish appropriate belief and
inject\slash acquire \textit{a priori} knowledge of the underlying environment
may be valuable directions to explore for automatic agent design. These
insights are further discussed with the case shown in the supplementary.



\begin{wraptable}{R}{0.4\linewidth}
    \addtolength{\tabcolsep}{-3pt}
    \centering
	\caption{Experiment results of VLM-based agents on WikiHow task set. The results
 are obtained with(\slash out) SoM and a different number of exemplars (Exem).  ``Claude'' denotes
 Claude-3-Opus.}
    \label{tab:vlm_rst}
    \begin{tabular}{lcccc}
        \toprule[1.5pt]
        Model & SoM? & \#Exems & Rwd & SR \\
        \midrule
        Claude & \ding{51} & 2 & 0.28 & 0.00 \\
        GPT-4V & \ding{51} & 2       & 0.90 & 10.00 \\
        GPT-4V & \ding{55} & 2       & 0.33 & 3.00 \\
        GPT-4V & \ding{51} & 0       & 0.70 & 5.00 \\
        \bottomrule[1.5pt]
    \end{tabular}
\end{wraptable}

\paragraph{Multi-turn prompting vs.\ single-turn prompting} We conducted
an experiment to investigate the impact of the prompting format for chat models. 
Multi-turn prompting is referred to as splitting the input exemplars into observation
and action and assigning observation the ``user'' role and action the
``assistant'' role. In contrast, single-turn prompting
is referred to as assigning all the input prompts (except the ``system'' part) the role
``user'' and expecting a 1-turn response from the LLM assistant. As depicted by the results
of GPT-3.5 in \cref{tab:t_llm_rst},
the single-turn format dramatically degrades the performance. We speculate that multi-turn
prompting combining the input-output logic with user-assistance distinction makes it clearer
for the model to learn the relation between inputs and outputs.

\subsection{Results of VLMs on fixed world set}

We evaluated the vision-based
agents on 40 tasks selected from the fixed world task set. As shown in \cref{tab:vlm_rst},  even the strongest 
VLM, GPT-4V, achieved only a 10.00\% success rate. Except for the reasons discussed in \cref{sub:op_expe}, it is 
additionally attributed to that current VLMs struggle to accurately ground on GUI screenshots. Notably, 
the SoM approach mitigates this challenge and improves VLM's performance by asking the VLMs to predict the 
index of a tagged screen element instead
of its precise coordinates. The result demonstrates there is significant room for improvement in current VLMs' ability to reason and correctly ground on the GUI screenshots.

\section{Conclusion}

Aiming at filling in the blank of a \xxx{} GUI interaction benchmark,
this work builds Mobile-Env as a comprehensive toolkit to simplify the construction of
\xxx{} benchmarks. Mobile-Env offers an isolated interactive environment and 
designs a flexible framework to produce reliable intermediate rewards and instructions.
With the help of Mobile-Env, two task sets are collected to draw insights into the performances
of LLM-based agents and provide truly reproducible and comparable results. The experiments
reveal the deficiency of current models in serving as GUI agents, underscoring the 
necessity to build more capable foundation models and more effective and robust agent 
frameworks augmented with knowledge of GUI understanding and operation.

\bibliographystyle{plainnat}
\bibliography{example_paper,custom,anthology}

\section*{Checklist}


\begin{enumerate}

\item For all authors...
\begin{enumerate}
  \item Do the main claims made in the abstract and introduction accurately reflect the paper's contributions and scope?
    \answerYes{}
  \item Did you describe the limitations of your work?
    \answerYes{The limitations and future work are discussed in the supplementary.}
  \item Did you discuss any potential negative societal impacts of your work?
    \answerYes{The potential societal impacts are discussed in the supplementary.}
  \item Have you read the ethics review guidelines and ensured that your paper conforms to them?
    \answerYes{}
\end{enumerate}

\item If you are including theoretical results...
\begin{enumerate}
  \item Did you state the full set of assumptions of all theoretical results?
    \answerNA{}
	\item Did you include complete proofs of all theoretical results?
    \answerNA{}
\end{enumerate}

\item If you ran experiments (e.g. for benchmarks)...
\begin{enumerate}
  \item Did you include the code, data, and instructions needed to reproduce the main experimental results (either in the supplemental material or as a URL)?
    \answerYes{The Mobile-Env code repository, WikiHow task set, and the crawled data of the WikiHow website are open-sourced at GitHub and Hugging Face. The links are attached to the main paper. The full prompts to reproduce the experiment results are represented in the supplementary as well.}
  \item Did you specify all the training details (e.g., data splits, hyperparameters, how they were chosen)?
    \answerNA{}
	\item Did you report error bars (e.g., with respect to the random seed after running experiments multiple times)?
    \answerNo{Duplicated experiments are not conducted owing to the limit of budget.}
	\item Did you include the total amount of compute and the type of resources used (e.g., type of GPUs, internal cluster, or cloud provider)?
    \answerYes{See the supplementary.}
\end{enumerate}

\item If you are using existing assets (e.g., code, data, models) or curating/releasing new assets...
\begin{enumerate}
  \item If your work uses existing assets, did you cite the creators?
    \answerYes{}
  \item Did you mention the license of the assets?
    \answerYes{The license is mentioned in the supplementary.}
  \item Did you include any new assets either in the supplemental material or as a URL?
    \answerYes{The links to the Mobile-Env code repository and WikiHow task set are attached in the main paper.}
  \item Did you discuss whether and how consent was obtained from people whose data you're using/curating?
    \answerNA{}
  \item Did you discuss whether the data you are using/curating contains personally identifiable information or offensive content?
    \answerNA{}
\end{enumerate}

\item If you used crowdsourcing or conducted research with human subjects...
\begin{enumerate}
  \item Did you include the full text of instructions given to participants and screenshots, if applicable?
    \answerNA{}
  \item Did you describe any potential participant risks, with links to Institutional Review Board (IRB) approvals, if applicable?
    \answerNA{}
  \item Did you include the estimated hourly wage paid to participants and the total amount spent on participant compensation?
    \answerNA{}
\end{enumerate}

\end{enumerate}

\appendix

\section{Distribution \& License}

The proposed Mobile-Env platform and WikiHow task set are open-sourced under
Apache-2.0 license. Mobile-Env is released on
\url{https://github.com/X-LANCE/Mobile-Env} and WikiHow task set is distributed
through \url{https://huggingface.co/datasets/X-LANCE/WikiHow-taskset}. Both
Mobile-Env and the task sets will be maintained continuously. The experiment
codes are open-sourced at \url{https://github.com/OpenDFM/mobile-env-expe}.

A non-exhaustive list of artifacts used in development of Mobile-Env and
WikiHow task set includes: AndroidEnv~\citep{DanielToyama2021_AndroidEnv},
mitmproxy\footnote{\url{https://mitmproxy.org}},
Pyserini~\citep{JimmyLin2021SIGIR_Pyserini},
WikiHow\footnote{\url{https://www.wikihow.com}}, \textit{etc.} These artifacts
are released under licenses Apache-2.0, MIT, CC-BY-NC-SA 3.0, \textit{etc.} A
non-exhaustive list of artifacts used in the experiments includes:
LLaMA-2~\citep{HugoTouvron2023_LLaMA2}, AgentLM~\citep{AohanZeng2023_AgentLM},
LLaMA-3~\citep{AIMeta2024_LLaMA3},
vLLM\footnote{\url{https://docs.vllm.ai/en/latest/index.html}},
Text-Generation-Inference\footnote{\url{https://github.com/huggingface/text-generation-inference}},
NLTK\footnote{\url{https://www.nltk.org/}},
EasyOCR\footnote{\url{https://www.jaided.ai/easyocr/}},
vision-ui\footnote{\url{https://github.com/Meituan-Dianping/vision-ui}},
\textit{etc}. They are released under licenses LLaMA 2, LLaMA 3, HFOILv1, MIT,
Apache-2.0, \textit{etc.} The authors claim that the usage completely obey the
licenses.

\section{Implementation details of Mobile-Env}
\label{sec:mobile_env}


\begin{figure}
    \centering
	\includegraphics[width=\linewidth]{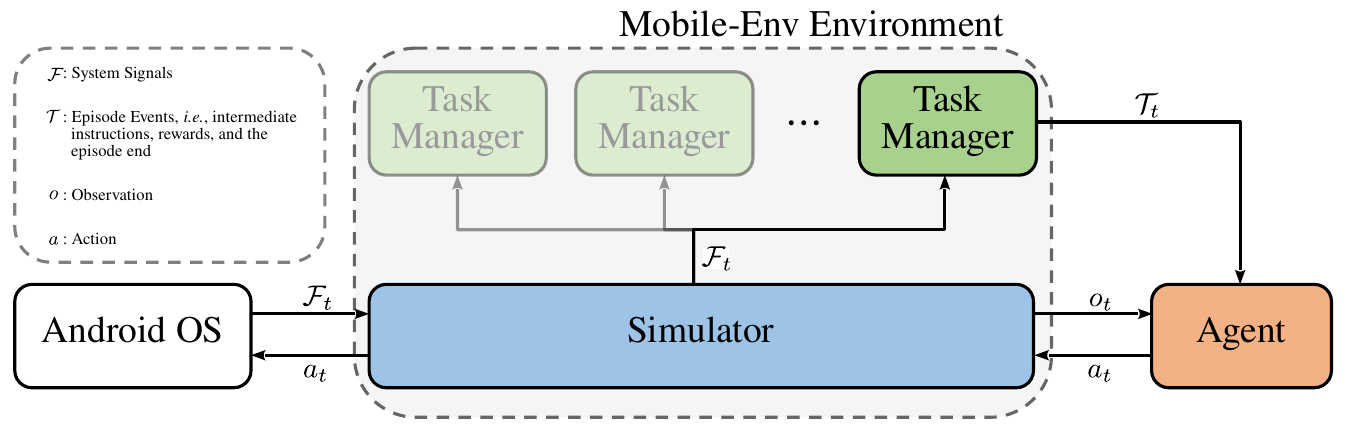}
	\caption{Main components of Mobile-Env. The simulator serves as a proxy to
		the Android OS, offers observations and accepts actions from the agent.
		The task manager estimates the task state through various types of
		system signals and produce episode events.  Multiple task managers can
		be combined with the same simulator to reduce overhead, and only one
	can be active for interaction at one time.}
    \label{fig:m_cmpnt}
\end{figure}

Mobile-Env is designed as a universal interaction platform hosting various apps
and different tasks to facilitate \xxx{} evaluation of LLM\slash VLM-based or
conventional GUI agents. A task is corresponding to a specific goal and several
configurations, which are defined by a task configuration file.  Based on
Mobile-Env, environment and task designers can easily build \xxx{} environments
and tasks. Then, agent designers can train and evaluate their GUI agents with
various environments and tasks held on Mobile-Env.

As shown in \cref{fig:m_cmpnt}, a runtime instance of Mobile-Env consists of
two types of components:
\begin{description}
	\item[Simulator] The simulator wraps Android
		Emulator\texttrademark{}\footnote{\url{https://developer.android.com/about}}
		and provides the agent with a suite of interaction interfaces with the
		Android\texttrademark{} OS. The simulator predefines basic observation
		space and action space.
	\item[Task manager] A task manager is corresponding to a specific task
		goal. During the launch of Mobile-Env instance, the task manager will
		be loaded by reading and parsing a task configuration file.  Then,
		during the episode, the task manager will produce intermediate
		instructions and rewards and indicate the end of episode (either
		failure or success) at the correct time according to the loaded task
		configuration.
\end{description}
At runtime, multiple task managers can be combined with one simulator to
prevent the overhead of multiple simulator instances. Note that there will be
only one task manager active for the current episode at one time.

During interaction, the agent will receive the observation from the simulator
and the task events (\textit{i.e.}, intermediate instructions, rewards, and the
episode end) from the active task manager. The action decision will then be
made accordingly and sent to the OS through the simulator.

\subsection{Mobile-Env simulator}

The simulator of Mobile-Env defines the basic observation space and action
space for GUI agent interaction. The observation includes both screenshot and
VH (View Hierarchy) XML. The basic raw action space consists of three types of
atomic actions, \verb|TOUCH|, \verb|LIFT|, and \verb|TEXT|. To keep the
interaction as efficient as possible, we adopt the
gRPC\footnote{\url{https://grpc.io/}, an open-source general-purpose Remote
Procedure Call (RPC) framework created by Google\texttrademark{}.} interfaces
of Android Emulator to obtain the screenshot and perform \verb|TOUCH| and
\verb|LIFT| actions. As there are not any gRPC methods to obtain VH and perform
\verb|TEXT|, we have to adopt ADB (Android Debug Bridge) to implement them,
which takes much longer time delay. A comparison of the efficiency of acquiring
the observation components and performing actions is provided in
\cref{tab:effc_comp}.

\begin{table}
	\centering
	\caption{Time delay of acquiring different types of observation and
	performing different actions}
	\label{tab:effc_comp}
	\begin{tabular}{cccc}
		\toprule[1.5pt]
		Item                  & Approach                      & Avg Time      & Time Std Dev \\
		\midrule
		Screenshot            & gRPC (\texttt{getScreenshot}) & 19.94 ms      & 21.47 ms     \\
		View Hierarchy        & ADB (\texttt{uiautomator})    & 2.53 s        & 1.90 s       \\
		\texttt{TOUCH} action & gRPC (\texttt{sendTouch})     & 419.50 $\mu$s & 64.71 $\mu$s \\
		\texttt{LIFT} action  & gPRC (\texttt{sendTouch})     & 412.30 $\mu$s & 84.18 $\mu$s \\
		\texttt{TEXT} action  & ADB (\texttt{input})          & 1.30 s        & 0.28 s       \\
		\bottomrule[1.5pt]
	\end{tabular}
\end{table}

\subsection{Certificate unpinning}
\label{sub:cert_pin}

As described in \S~3.2.1 in the main paper, certificate pinning prevents
real-world apps from trusting the SSL certificates issued by the replay proxy.
Thus, to establish truly controlled environments, we sought help from reverse
engineering and developed three solutions. 
\begin{enumerate*}[label=\arabic*)]
	\item Injecting the proxy's certificates into a writable system image and
		making it mock a built-in system certificate.
	\item Replacing the verifier used by the app at runtime via runtime
		instrument tool.
	\item Repackaging the APK (installation package of the app) to remove its
		configurations regarding certificate pinning.
\end{enumerate*}
We validated the solutions on a wide series of common-use apps and affirm that
most apps can be dealt with by at least one solution. 


\subsection{Task state estimation}

\begin{wrapfigure}{R}{0.4\linewidth}
    \centering
    \includegraphics[width=\linewidth]{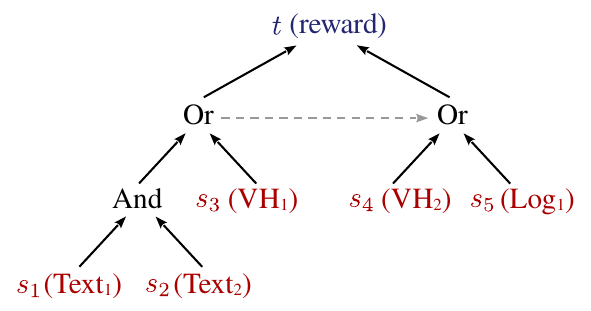}
	\caption{A demonstration of combining different types of system signals in
	a tree form.  {\color{dullred} ``Text''}, {\color{dullred} ``VH''}, and
{\color{dullred} ``Log''} indicate different types of signals. }
    \label{fig:demo_evt_tr}
\end{wrapfigure}

Mobile-Env leverages five primary types of signals to enable reliable task
state estimation and produce intermediate instructions and rewards and episode
ends accordingly at the right time.
\begin{description}
	\item[Screen text] Mobile-Env detects or recognizes some texts matching a
		specific pattern from the screenshot. An external OCR
		system is required.
	\item[Screen icon] Mobile-Env detects or recognizes some icons matching a
		specific pattern from the screenshot. The pattern can be specified
		through an icon class or a reference image. An external icon
		recognition system is needed.
	\item[View hierarchy] Mobile-Env checks if an expected node exists in VH of
		the current screen. VH is acquired through ADB \verb|uiautomator|.
	\item[System log] Mobile-Env monitors the system log for some lines
		matching a specific pattern. System log is acquired through ADB
		\verb|logcat|. 
	\item[Response to human user (RHU)] Mobile-Env listens to the response
		generated by the agent and checks if it matches the groundtruth answer.
\end{description}

\begin{figure}[t]
	\centering
	\subfigure[The instruction is ``Check my TOTP for GitHub and tell me the
		code''. As the TOTP is a time-based dynamic passcode, the evaluator
		will compare the answer from the agent with the real-time passcode
		shown in the authenticator. The text is detected according to a regex
		(regular expression).  The VH node is seletect via a customized CSS
	selector (used to filter XML
nodes).]{\includegraphics[width=0.95\linewidth]{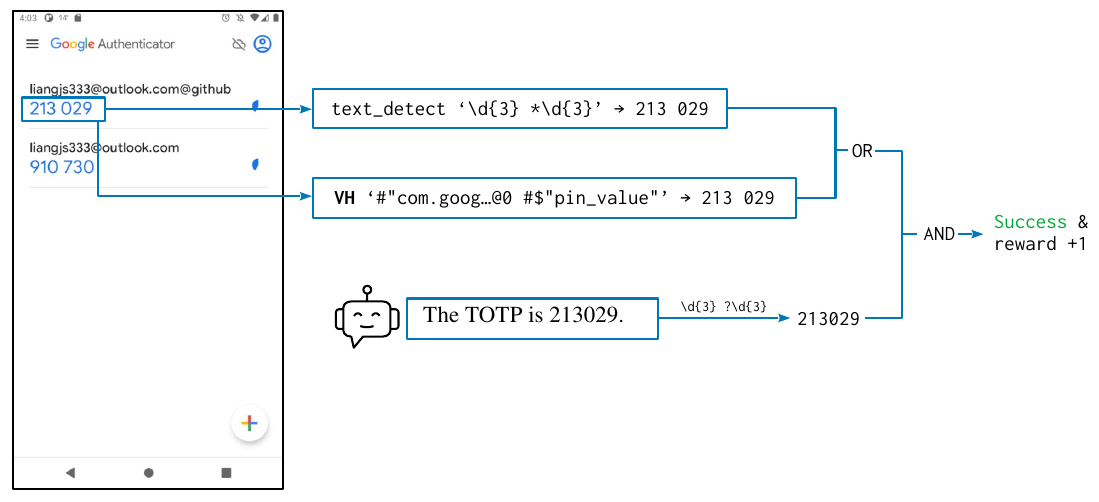}}

	\subfigure[The instruction is ``Turn off my alarm in the morning if it will
		rain tomorrow, otherwise leave it on''. The agent must first check the
		weather of tomorrow through the weather app on the phone before it
		switches the alarm clock on or off
	accordingly.]{\includegraphics[width=0.76\linewidth]{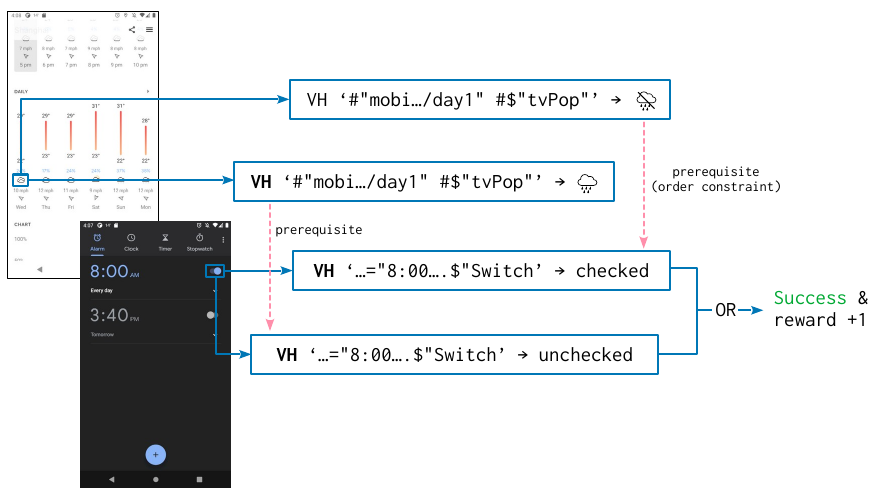}}
	\caption{Examples of concrete task state estimation and evaluation via Mobile-Env}
	\label{fig:eval_eg}
\end{figure}

Mobile-Env further introduces three strategies for combining these signals to
enhance the flexibility of state estimation: conjunctive evaluation,
disjunctive evaluation, and order constraint. Mobile-Env proposes to represent
these combination strategies in a tree form. As shown in
\cref{fig:demo_evt_tr}, we design \texttt{AND} and \texttt{OR} nodes to express
conjunctive and disjunctive evaluations, respectively. Skip connections (the
\textcolor{linegrey}{gray} dashed line) are used to express the order
constraints between different signals (events). In this form, the state
estimation strategy can be expressed flexibly and clearly in the task
configuration file.  Two concrete examples of state estimation and reward
generation via Mobile-Env are illustrated in \cref{fig:eval_eg}.

\subsection{Miscellaneous auxiliary tools}
\label{sub:misc_tool}

\paragraph{Annotation tool for human demonstration} Touch and typing are close
to human behaviors and are suitable to apply behavior cloning approaches.
Based on Mobile-Env, an annotation tool is developed to collect human
demonstrations for convenience. The annotation tool offers a straightforward
web interface to collect the human actions. The annotators can directly click
on the shown screenshot to mock touching on the mobile screen, or click on a
button on the webpage to directly type a word from a shrunk vocabulary. The
actions, observations, and task information will be recorded for behavior
cloning.

\paragraph{Task definition template toolkit} A simple suite of tools is
developed to auto-generate task configuration files from templates. Such a task
configuration template declares a group of value slots. The value slots can be
instantiated according to a simple instantiation configuration file.  Besides,
the configuration instantiated from a single template can be combined
sequentially to form a ``multi-stage'' task definition. The fixed world WikiHow
task set is built with the help of this toolkit.

\section{Details about the task sets}
\label{sec:wikihow}

\subsection{Instructions of open world set}

The full list of (the first) instruction in the open world set is listed in
\cref{tab:open_wd_set_inst}.

\input{open-set-instructions.tex}

\subsection{App data crawling and replay for WikiHow task set}
\label{subsec:app_dat_cllct}

In order to establish a truly reproducible and comparable benchmark, we need to
crawl and store the varying app data of WikiHow to build the fixed world set.
The app data will be replayed at runtime to mimic the real scenario.

WikiHow app comprises two types of pages: static pages and dynamic pages. The
static pages are referred to as the pages independent of user
inputs\footnote{Static pages may contain contents that vary according to
	spatio-temporal conditions as well, \textit{e.g.}, dynamic recommendations,
statistics for visits, and ratings. However, these contents do not depend on
instant user inputs, and thus are considered ``static'' resources.} such as the
home page, articles, author information, \textit{etc}. The dynamic pages are
the pages sensitive to user inputs, such as the pages of search results.  App
data for static pages are directly collected from WikiHow
website\footnote{\url{https://www.wikihow.com/Main-Page}}. As observed that the
pages in the app are different from the pages in the browser, we checked the
HTTP request headers of all the types of pages and mocked the app requests to
obtain the correct contents. Considering that the user always explores the app
from the home page, the crawler also starts from the home page and accesses the
links in breadth-first order. For simplicity, the links to the outside of
WikiHow are omitted. All the links including all types of media are requested
and stored for a perfect replay. Finally, 856,045 resources are dumped, in
which there are 107,448 distinct pages. These website resources totally occupy
about 88 GiB. In contrast to the static resources, it is nearly impossible to
traverse all the probable combinations of keywords to dump the pages of search
results.  Consequently, an open-source information retrieval engine,
Pyserini~\citep{JimmyLin2021SIGIR_Pyserini}, is adopted following
\citet{ShunyuYao2022_WebShop} to mimic the WikiHow search engine. Then the
synthetic search pages are built dynamically during the replay upon the
template of the search result page in WikiHow app.


The crawled data are tagged with the requested URL paths. Each response is
dumped in an individual file. Both HTTP headers and raw payload are saved. The
flow data can be replayed through
mitmproxy\footnote{\url{https://mitmproxy.org/}}.  The saved static resources
are directly read and replayed with the timestamps refreshed. The search result
page is crafted with the results from Pyserini
engine~\citep{JimmyLin2021SIGIR_Pyserini} and the page template. Pyserini
builds the indices of the articles from the article titles.  The page template
is built from a real search result page in WikiHow app by making the search
result list a filling slot.

\subsection{Automatic task generation for WikiHow task set}
\label{subsec:t_gnrt_instrct_rwrt}

By thoroughly exploring WikiHow app, 10 basic operations on WikiHow are
summarized as:
\begin{itemize}
	\item Search for an article,
	\item Access an article from the current page,
	\item Access the about page from article page,
	\item Access author information page from article page,
	\item Access category content page from article page,
	\item Check reference list,
	\item Rate yes\slash no for article,
	\item Share article,
	\item Bookmark article,
	\item Check the bookmarks.
\end{itemize}
Based on this summary, a suite of ``single-stage'' sub-task configuration
templates is designed:
\begin{itemize}
	\item \verb|home2search|, instructing to search for an article from the
		home page.
	\item \verb|search2article|, \verb|author2article|, \&
		\verb|category2article|, instructing to access an article from search
		result page, author information page, and category content page,
		respectively.
	\item \verb|article2about|, instructing to access the about page from
		article page.
	\item \verb|article2author|, instructing to access author information page
		from article page.
	\item \verb|article2category|, instructing to access category content page
		from article page.
	\item \verb|article2reference|, instructing to check reference list on
		article page.
	\item \verb|article2rate_no|, instructing to rate no for article
	\item \verb|article2rate_yes|, instructing to rate yes for article
	\item \verb|article2share|, instructing to share article
	\item \verb|article2bookmark|, instructing to bookmark article and then
		check the bookmarks.
\end{itemize}
among which, \verb|article2reference|, \verb|article2rate_no|,
\verb|article2rate_yes|, \verb|article2share|, and \verb|article2bookmark| are
used to form in-page tasks. In addition, we explore two open-source QA datasets
based on WikiHow,
wikihow-lists\footnote{\url{https://huggingface.co/datasets/b-mc2/wikihow_lists}}
and
WikiHowNFQA\footnote{\url{https://huggingface.co/datasets/Lurunchik/WikiHowNFQA}},
to construct QA tasks in WikiHow task set. wikihow-lists comprises three types
of question:
\begin{itemize}
	\item \verb|stepped_summary|, expecting a list of key steps summarized from
		the current article.
	\item \verb|ingredients|, expecting a list of ingredients for a recipe
		extracted from the current article.
	\item \verb|needed_items|, expecting a list of needed items for an activity
		extracted from the current article.
\end{itemize}
WikiHowNFQA posts free-form summarization questions related to WikiHow
articles. Based on these question-answer pairs, we additionally craft 4
sub-task configuration templates:
\begin{itemize}
	\item \verb|article2steps|, crafted from \verb|stepped_summary|s
	\item \verb|article2ingredientes|, crafted from \verb|ingredients|s
	\item \verb|article2needed_items|, crafted from \verb|needed_items|s
	\item \verb|article2summary|, crafted from WikiHowNFQA questions
\end{itemize}
Each template contains a group of filling slots expecting some keywords like
article title, author name, question, and groundtruth answer. Then these
keywords are sampled from the crawled app data or from the two QA datasets to
instantiate the templates. Subsequently, the instantiated templates are
concatenated into multi-stage task definitions under the constraint that the
target page\slash element\slash answer (the part after ``\verb|2|'',
\textit{e.g.}, ``\verb|share|'' from ``\verb|article2share|'') is directly
on\slash referenced by the current page (the part before ``\verb|2|'',
\textit{e.g.}, ``\verb|article|'' from ``\verb|article2share|''). Finally, we
obtained the task set of 150 multistage tasks in which there are 2.68
single-stage sub-tasks averagely.


\begin{figure}[t]
	\centering
	\subfigure[Examples of human-rewritten
	instruction \label{fig:hum_inst}]{\includegraphics[width=0.45\linewidth]{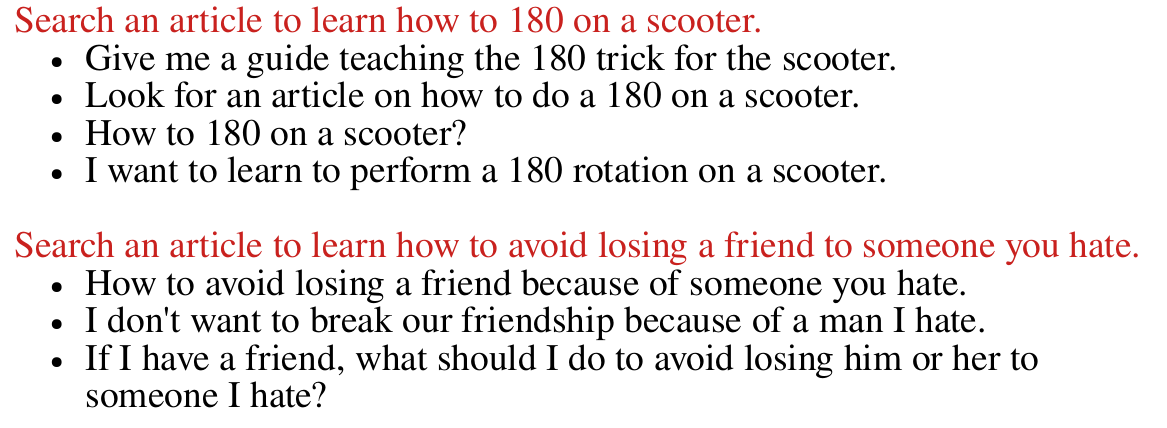}}
	~
	\subfigure[Examples of summarized public instruction
	pattern \label{fig:hum_inst_pat}]{\includegraphics[width=0.45\linewidth]{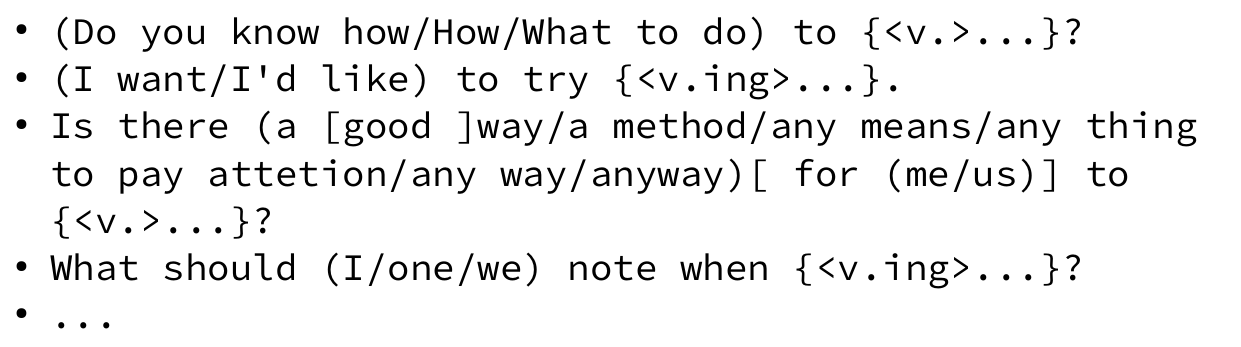}}
	\caption{Examples of human-rewritten instructions (Left) and public
	instruction patterns (Right)}
	\label{fig:instrct_pttn}
\end{figure}

\subsection{Instruction rewriting \& dynamic replacement for WikiHow task set}
\label{sub:rewriting}

The template-generated task definitions of WikiHow are homogeneous and
straightforward, directly containing the target article name to search or
access. These instructions are too simple for LLMs to understand. In order to
establish a more challenging benchmark and evaluate the capacity of LLM for
instruction understanding, the authors invited several colleagues and leveraged
ChatGPT to rewrite the stereotyped instructions. Each person created 2--3 new
expressions for each instruction. To further improve diversity, the authors
inspected all the rewritten instructions and summarized several public
expression patterns that are easy to transfer to different topics.  Several
examples of rewritten instruction and public pattern are depicted in
\cref{fig:instrct_pttn}.  The corresponding rewritten instructions and public
patterns are randomly sampled to replace the original ones at runtime.  The
different branches in pattern (\textit{e.g.}, {\tt (I want\slash I'd like)})
are randomly selected and the keyword in slots (\textit{e.g.},
\verb|{<v.ing>...}|) are transformed into the correct POS (part-of-speech).
POS recognition and transformation are implemented with the help of
NLTK\footnote{\url{https://www.nltk.org}} and
lemminflect\footnote{\url{https://github.com/bjascob/LemmInflect}}.


%

\section{Details about LLM-based agents and experiments}
\label{sec:llm_agent}


\subsection{LLM access and compute resources}
\label{sub:llm_intro}

We tested the closed-source text LLMs including GPT-3.5 (gpt-3.5-turbo-1106 \&
gpt-3.5-turbo-instruct), GPT-4\footnote{\url{https://platform.openai.com}}
(gpt-4-1106-preview), and
Claude-3\footnote{\url{https://www.anthropic.com/claude}}
(claude-3-opus-20240229).  The capable closed-source VLMs are also tested:
GPT-4V (gpt-4-1106-vision-preview) and Claude-3 (claude-3-opus-20240229). These
models are accessed through the online APIs (Application Programming
Interfaces) provided by OpenAI and Anthropic.

We tested open-source LLMs including LLaMA 2~\citep{HugoTouvron2023_LLaMA2}
(llama-2-chat-7b, llama-2-chat-13b, and llama-2-chat-70b),
AgentLM~\citep{AohanZeng2023_AgentLM} (agentlm-7b, agentlm-13b, and
agentlm-70b), and LLaMA 3~\citep{AIMeta2024_LLaMA3} (llama-3-8b-instruct \&
llama-3-70b-instruct). These models are deployed on NVIDIA A10 Tensor Core
clusters with bfloat16 with the help of Hugging Face
Text-Generation-Inference\footnote{\url{https://github.com/huggingface/text-generation-inference}}
and vLLM\footnote{\url{https://docs.vllm.ai/en/latest/index.html}}. 7B (8B),
13B, and 70B models require 1, 2, and 8 A10 GPUs with 24 GiB memory,
respectively.

In our experiments, Android Emulator runs on a Linux host with KVM
(Kernel-based Virtual Machine) enabled to conduct efficient enough emulation.


\begin{table}[t]
	\begin{minipage}{0.49\linewidth}
		\centering
		\caption{Mapping from VH node classes to HTML tags}
		\label{tab:vh_n_html_t}
		\begin{tabular}{cc}
			\toprule[1.5pt]
			VH node class & HTML tag class \\
			\midrule
			\verb|*TextView| &  \verb|p| \\
			\verb|*Button| &    \verb|button| \\
			\verb|*MenuItemView| &   \verb|button| \\
			\verb|*ImageView| & \verb|img| \\
			\verb|*IconView| &  \verb|img| \\
			\verb|*Image| &  \verb|img| \\
			\verb|*EditText| &  \verb|input| (\verb|type="text"|) \\
			others & \verb|div| \\
			\bottomrule[1.5pt]
		\end{tabular}
	\end{minipage}
	~
	\begin{minipage}{0.49\linewidth}
		\centering
		\caption{Mapping from VH properties to HTML attributes. The
			\texttt{resource-id} property usually consists of three parts: the
			package name, the element class, and the element name. Only the
			element name is used for the converted \texttt{class} attribute,
			which follows \citet{BryanWang2022_ConversationMobileUI_LLM}.}
		\label{tab:vh_prpt_html_attrb}
		\begin{tabular}{cc}
			\toprule[1.5pt]
			VH property & HTML attribute \\
			\midrule
			\verb|resource-id| &    \verb|class| \\
			\verb|content-desc| &   \verb|alt| \\
			\bottomrule[1.5pt]
		\end{tabular}
	\end{minipage}
\end{table}

\begin{figure}[t]
	\centering
	\includegraphics[width=\linewidth]{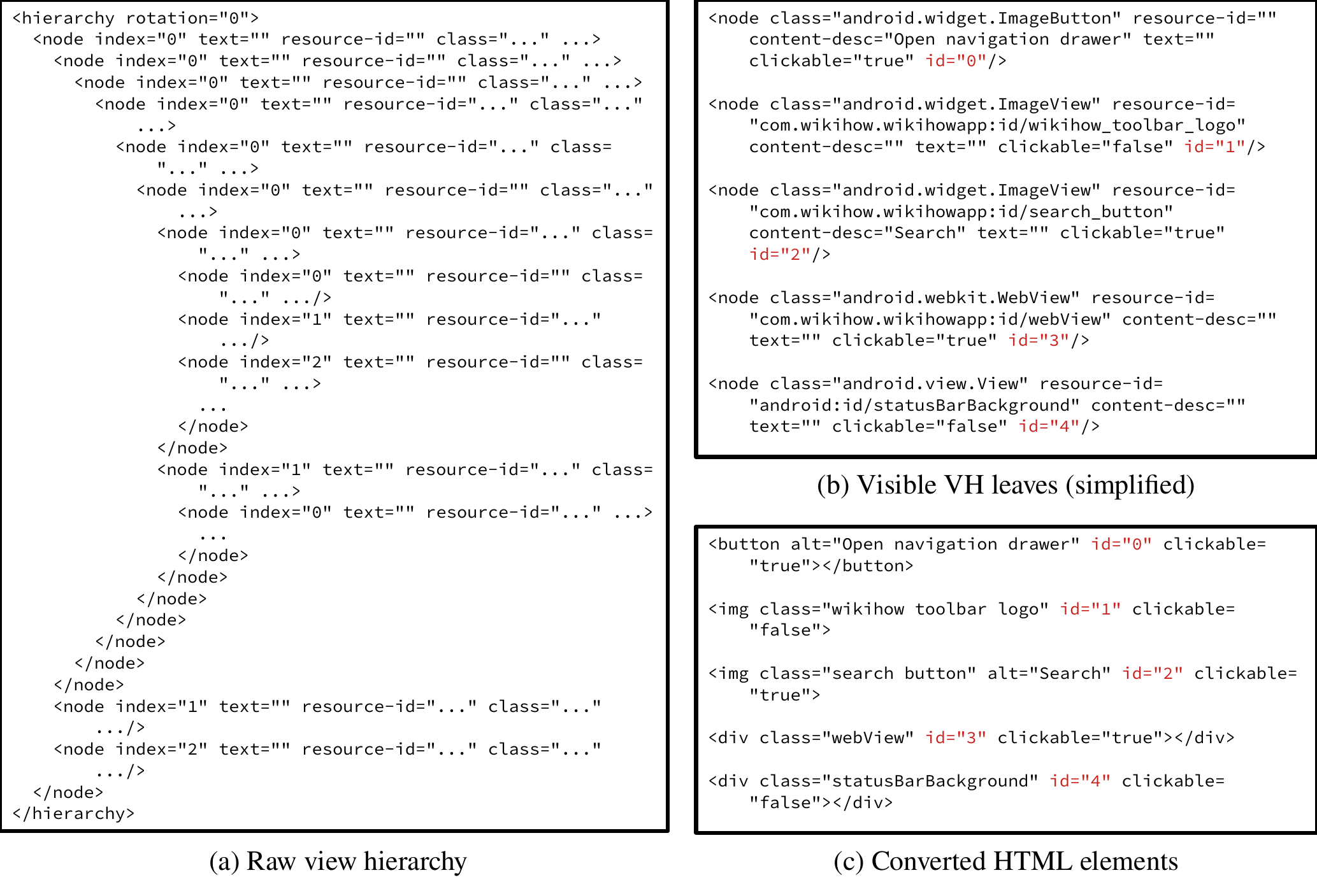}
	\caption{Example of conversion from VH into HTML. Fig.~(a) depicts the raw
		view hierarchy extracted from the OS. Fig.~(b) depicts the visible leaf
		nodes from Fig.~(a). Additional \texttt{id} property is appended.
		Fig.~(c) depicts the converted HTML elements from Fig.~(b) according
		to the conversion rule.}
	\label{fig:vh_to_html}
\end{figure}

\subsection{Observations for LLM\slash VLM-based agents}
\label{sub:vh_to_html}

The text LLMs are fed with a simplified HTML format derived from the VH
component in the raw observation following
\citet{BryanWang2022_ConversationMobileUI_LLM}. In the HTML, only the visible
leaf nodes in VH are preserved. The VH classes and essential properties are
converted into HTML tags and attributes. \cref{tab:vh_n_html_t} depicts the
mapping rule from classes of VH nodes to HTML tags.
\cref{tab:vh_prpt_html_attrb} depicts how the VH node properties are mapped to
the HTML element attributes.  An example of conversion from VH into HTML is
illustrated in \cref{fig:vh_to_html}.

The VLMs are fed with screenshots applied Set-of-Marks
(SoM)~\citep{JianweiYang2023_SoM}. We adopt
EasyOCR\footnote{\url{https://www.jaided.ai/easyocr/}} to detect the screen
texts and
vision-ui\footnote{\url{https://github.com/Meituan-Dianping/vision-ui}} to
detect the screen icons and mark the detected elements using numeric tags.

\begin{table}[t]
	\centering
	\caption{Actions spaces for different types of LLMs\slash VLMs}
	\label{tab:act_spc}
	\begin{tabular}{m{1cm}<{\centering}ll}
		\toprule[1.5pt]
		Model Type & Action & Description \\
		\midrule
		\multirow{6}{*}{\rotatebox[origin=c]{90}{\parbox[c]{1.5cm}{\centering\noindent Text LLM}}}
        & {\tt CLICK(element\_id)}       & Click on an element.                                                  \\
        & {\tt LONG\_CLICK(element\_id)} & Long click on an element.                                             \\
        & {\tt INPUT(element\_id, text)} & Click on an element and input texts into it.                          \\
        & {\tt SCROLL(direction)}        & Scroll the page (\textit{i.e.} slide towards the opposite direction). \\
        & {\tt ANSWER(text)}             & Generate an answer response to human user.                            \\
        & {\tt GOBACK}                   & Go back to the last page.                                             \\
		\midrule
		\multirow{6}{*}{\rotatebox[origin=c]{90}{\parbox[c]{1.5cm}{\centering\noindent VLM w/ SoM}}}
        & {\tt TAP(element\_id)}         & Tap an element.                                                       \\
        & {\tt LONG\_TAP(element\_id)}   & Long tap an element.                                                  \\
        & {\tt TYPE(text)}               & Directly type some texts.                                             \\
        & {\tt SLIDE(direction)}         & Slide the ``finger'' towards a direction.                             \\
        & {\tt ANSWER(text)}             & Generate an answer response to human user.                            \\
        & {\tt GOBACK}                   & Go back to the last page.                                             \\
		\midrule
		\multirow{6}{*}{\rotatebox[origin=c]{90}{\parbox[c]{1.5cm}{\centering\noindent VLM w/o SoM}}}
        & {\tt TAP(x, y)}                & Tap a pixel.                                                          \\
        & {\tt LONG\_TAP(x, y)}          & Long tap a pixel.                                                     \\
        & {\tt TYPE(text)}               & Directly type some texts.                                             \\
        & {\tt SLIDE(x0, y0, x1, y1)}    & Slide the ``finger'' from {\tt (x0, y0)} to {\tt (x1, y1)}.           \\
        & {\tt ANSWER(text)}             & Generate an answer response to human user.                            \\
        & {\tt GOBACK}                   & Go back to the last page.                                             \\
		\bottomrule[1.5pt]
	\end{tabular}
\end{table}

\begin{figure}[tb]
	\centering
	\subfigure[Sliding trajectory of \texttt{SCROLL(DOWN)} (sliding
	up)]{\includegraphics[width=0.37\linewidth]{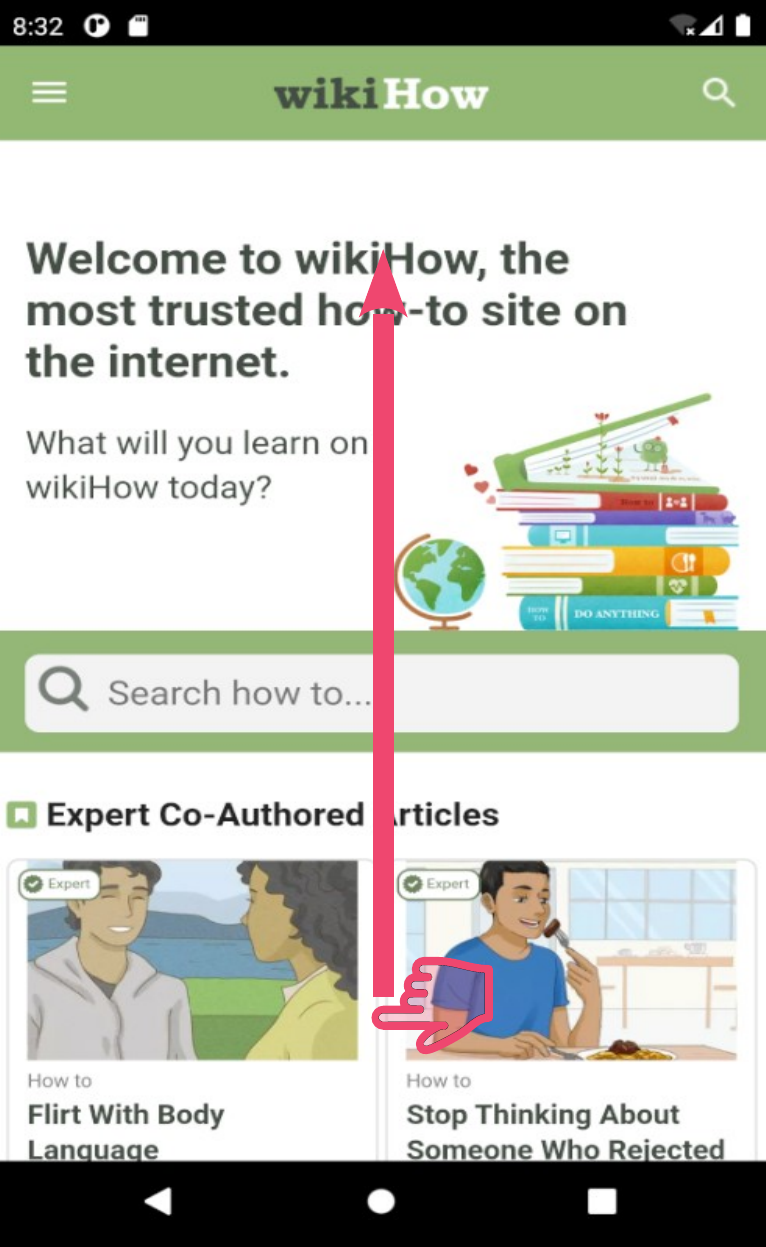}}
	~
	\subfigure[Sliding trajectory of \texttt{SCROLL(RIGHT)} (sliding
	left)]{\includegraphics[width=0.37\linewidth]{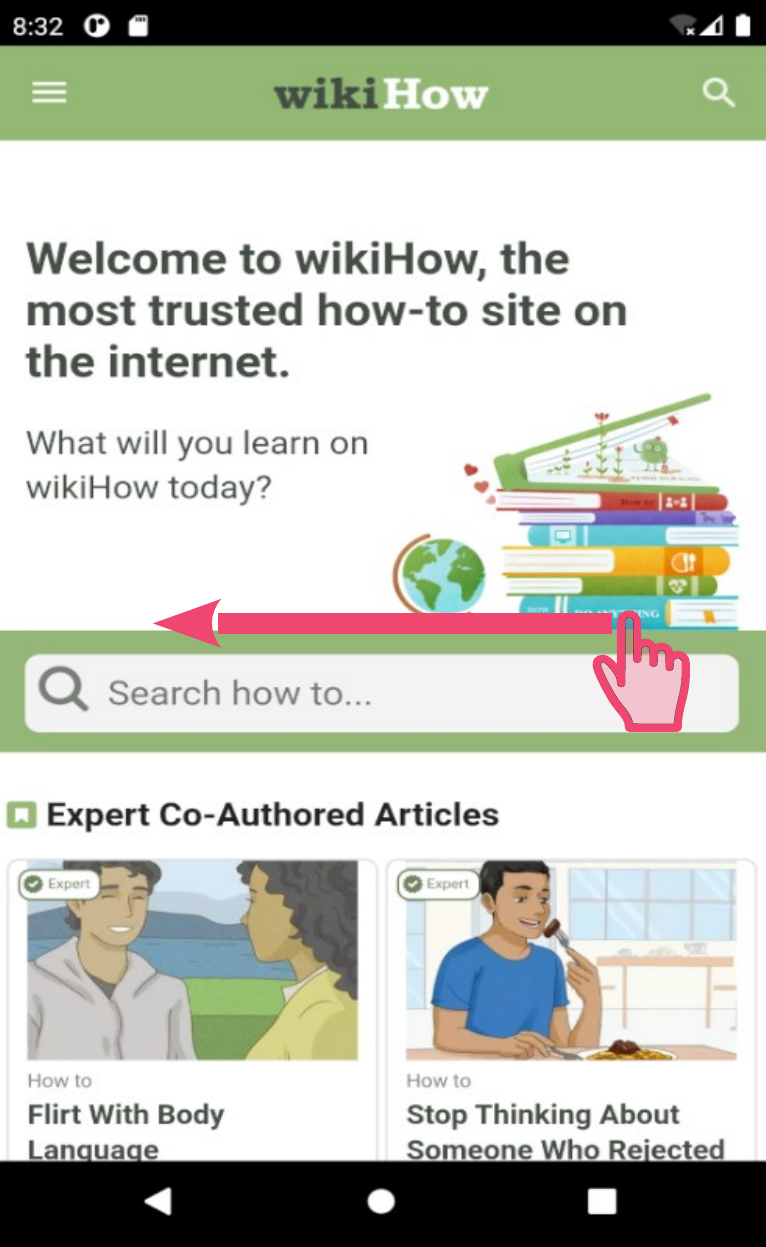}}
	\caption{Demonstration of sliding trajectories translated from
	\texttt{SCROLL}. The starting point is at 20\% of the screen height\slash
width, while the terminating point is at 80\%.}
	\label{fig:slide_trjct}
\end{figure}

\subsection{Action spaces for LLM\slash VLM-based agents}
\label{sub:vh_act_to_basic}

The designed action spaces for models are shown in \cref{tab:act_spc}.

These actions (\textit{i.e.}, \verb|CLICK|, \verb|LONG_CLICK|, \verb|INPUT|,
\verb|SROLL|, \verb|TAP|, \verb|LONG_TAP|, \verb|TYPE|, and \verb|SLIDE|) are
supposed to be translated into basic actions (\textit{i.e.}, \verb|TOUCH|,
\verb|LIFT|, and \verb|TEXT|) to be executed in Mobile-Env. Particularly,
\verb|CLICK|, \verb|LONG_CLICK|, \verb|TAP|, and \verb|LONG_TAP| are translated
into a sequence of \verb|TOUCH|s appended with a \verb|LIFT|, \textit{e.g.},
\begin{equation}
	\mathtt{CLICK}(\mathtt{eid}) \mapsto \underbrace{\mathtt{TOUCH}(c), \cdots, \mathtt{TOUCH}(c)}_{\text{$n$ \texttt{TOUCH}s}}, \mathtt{LIFT}.
	\label{eqn:click_trslt}
\end{equation}
Here \verb|eid| is the integer element id and $c$ is the center point of the
element bounding box. $n$ is set to 3 for \verb|CLICK| and \verb|TAP| and 10
for \verb|LONG_CLICK| and \verb|LONG_TAP| in our experiments.  \verb|TYPE| is
translated into a sequence of \verb|TEXT|s, and a keyboard action ``Enter'':
\begin{equation}
	\mathtt{TYPE}(\mathtt{text}) \mapsto \mathtt{TEXT}(\mathtt{tkn}_1), \cdots, \mathtt{TEXT}(\mathtt{tkn}_l), \mbox{Enter},
	\label{eqn:input_trslt}
\end{equation}
where $\mathtt{tkn}_i$ denotes the $i$-th token from the input \verb|text|. The
keyboard action ``Enter'' is a special action beyond the basic actions and
indicates typing a ``carriage return'' into the OS.  \verb|INPUT| is translated
into a \verb|CLICK| followed by a \verb|TYPE|.  \verb|SCROLL| and \verb|SLIDE|
are translated into $m$ consecutive \verb|TOUCH|s followed by a \verb|LIFT|,
where $m$ depends on the sliding distance. These \verb|TOUCH|s form a
consecutive sliding of the ``finger'' on the screen.  \cref{fig:slide_trjct}
demonstrates the translated sliding trajectories of typical \verb|SCROLL|s.

\subsection{Hyperparameter selection}
\label{sub:hypparam_slct}

It is found that the maximum length of 2 exemplars can just fit in the 4K-token
context of the open-source models (LLaMA 2, AgentLM, and LLaMA 3). Thus, 2-shot
in-context learning (ICL) is adopted for a fair experiment setting. A relative
low sampling temperature 0.1 is selected to balance the generation of
structured action representations and free-form reasoning languages, also to
balance the reproducibility and exploration ability of models. Before formal
experiments, the authors invited several colleagues to complete the designed
tasks. This is implemented by simply replacing the revocation of LLM with
keyboard inputting. Text observations are simply printed to the terminal, and
screenshot observations are shown in a separate window. Their feedback reveals
that the aforementioned observation \& action spaces are sufficient to complete
the tasks, and most of the tasks can end in 5 steps.  Therefore, the episodes
with more than 15 steps are considered failures during formal evaluation.

\subsection{Prompts for LLM\slash VLM-based agents}
\label{sub:prompts}


The complete prompts for LLMs and VLMs are provided in
\cref{tab:text_llm_prmpt} and \cref{tab:vis_llm_prmpt}.

\input{chat_prompt}

\input{mm_prompt}

\subsection{More analyses on open world set}

\begin{wraptable}{R}{0.57\linewidth}
	\centering
	\caption{Ablation study regarding the difficulty of ``starting from
		homepage''. ``App'' indicates the involved app is open at the task
		beginning, while ``Home'' indicates the app is not open and the agent
		has to complete the task by open the correct app firstly. However, the
		agents struggle to open the correct app when facing the Android home
	page, and thus obtain worse results.}
	\label{tab:abl_appop_vs_scratch}
	\begin{tabular}{lcccc}
		\toprule[1.5pt]
		\multirow{2}{*}{Model} & \multicolumn{2}{c}{SR} & \multicolumn{2}{c}{Rwd} \\
		\cmidrule(lr){2-3}\cmidrule(lr){4-5}
                               & App                    & Home                     & App    & Home   \\
		\midrule
		Claude-3-Opus          & 14.70                  & 8.83                     & 0.18 & 0.09 \\
		GPT-4                  & 38.23                  & 23.53                    & 0.41 & 0.26 \\
		AgentLM-70B            & 8.82                   & 0.00                     & 0.09 & 0.00 \\
		LLaMA-3-70B            & 20.59                  & 12.50                    & 0.21 & 0.15 \\
		\bottomrule[1.5pt]
	\end{tabular}
\end{wraptable}

\paragraph{Agents fail to open correct apps from the Android homepage.} To
reduce difficulty, the environments are launched with the involved app open
before the agent's execution. Wondering the raw performance of the agents, we
further choose the first 34 tasks in \cref{tab:open_wd_set_inst} (the tasks
involving various apps beyond WikiHow, Douban, IMDB, and Ele.me) and conduct
ablation study by making the agent start the task from the homepage. The
results are shown in \cref{tab:abl_appop_vs_scratch}. In this setting, the
involved app is not open in advance, and the agent has to open the correct app
by itself. This seems to be hard for most models. The difficulty mainly lies on
the way to open the app menu. The correct approach is to slide up on the home
page, but the agents often fail to predict this action. The agents may also be
misled, \textit{e.g.}, they always mistake the icon of Google ({\tt <img
class="g icon" id="6" clickable="true">}) for the icon of Gmail and approach
failure.

There are other difficulties revealed by our experiments that hinder the LLM
and VLM agents from completing tasks. For example, VH is not available on
several sensitive pages, \textit{e.g.}, the page showing the version of Android
OS. Thus, such tasks can only be completed by visual models like GPT-4V. The
safety barrier of the models can also lead to failed execution, \textit{e.g.},
Claude-3-Opus refuses to check the TOTP for the human user. Inexplicable
strange activities also exist, such as GPT-4V always existing the WikiHow app
at once when encountering tasks on WikiHow.

\subsection{More analyses \& case study on fixed world set}
\label{sub:c_std}

\begin{figure}[t]
	\centering
	\includegraphics[width=\linewidth]{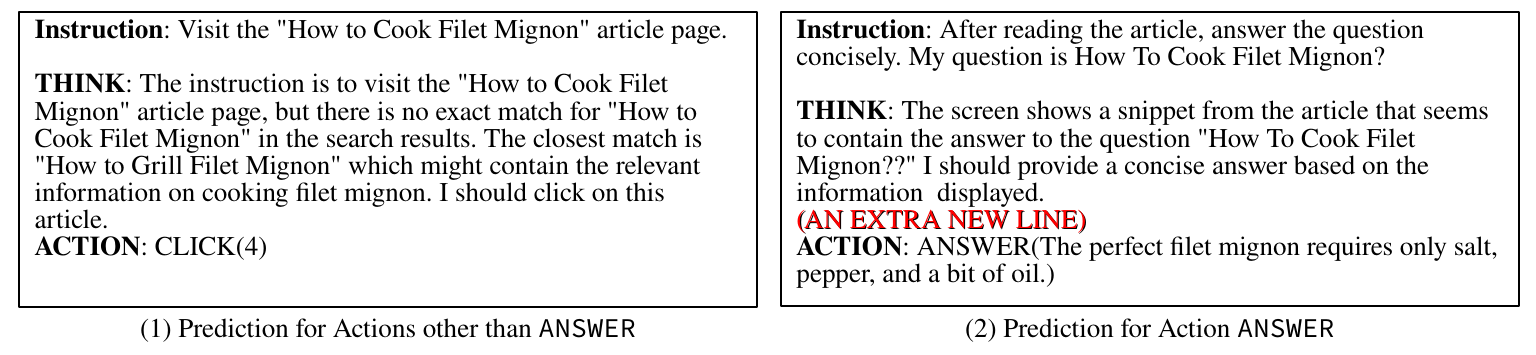}
	\caption{Different behavior of GPT-4 when predicting different types of
	action. When outputting \texttt{ANSWER} action, the model is prone to
predict an extra ``new line'' between the ``THOUGHT'' and ``ACTION'' lines.}
	\label{fig:gpt4_qa_nl}
\end{figure}

\begin{figure}[t]
	\centering
	\subfigure[Chat model outperforms instruct model on in-page
	tasks.]{\includegraphics[width=0.45\linewidth]{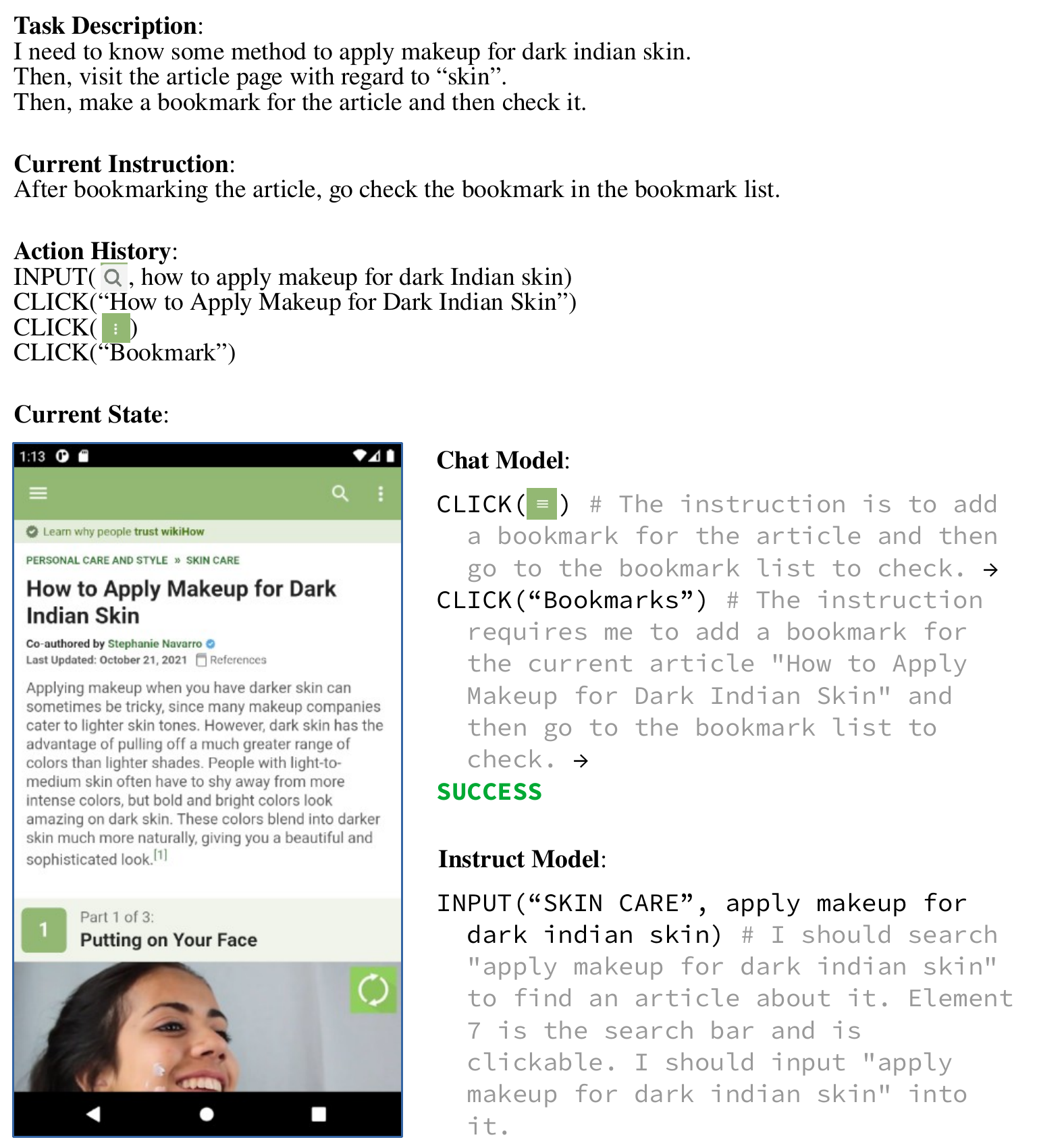}}
	~
	\subfigure[Instruct model performs better on cross-page
	tasks.]{\includegraphics[width=0.45\linewidth]{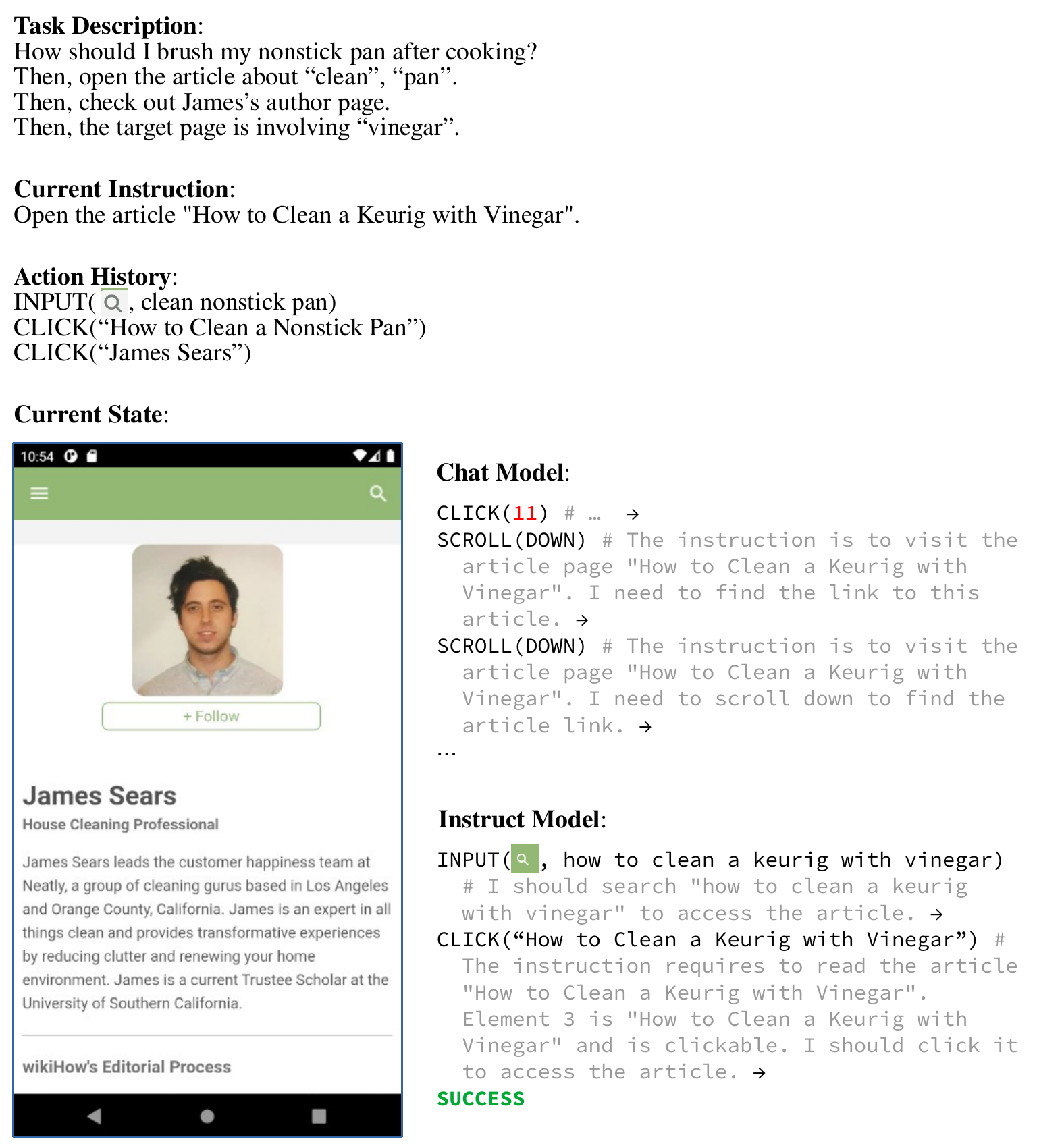}}
	\caption{Capability difference of chat model and instruct model. For
	clarity, the element ids are replaced with its text or icon in the figure.
Non-existing element id is marked with red. Thought texts are rendered in gray
and separated with action text by ``\#''.}
	\label{fig:chat_vs_instruct}
\end{figure}

\paragraph{Failure of GPT-4 on QA tasks} GPT-4 mysteriously predicts an extra
``new line'' between its thought and \verb|ANSWER| action, which violates the
output format and leads to failure. A group of examples is shown in
\cref{fig:gpt4_qa_nl}. As discussed in \S~5.2 in the main paper, this
phenomenon has become a significant bottleneck of the capable models'
performance.

\paragraph{Capability difference of chat and instruct models} The chat model
(GPT-3.5) and instruct model (GPT-3.5-instruct) do not show significant
disparity in the average performance (37.33\% SR vs. 36.67\% SR), however,
demonstrate the opposite advantages of completing in-page and cross-page tasks
(38.98\% SR on cross-page tasks and 49.02\% SR on in-page tasks of GPT-3.5 \&
44.07\% SR on cross-page tasks and 35.29\% SR on in-page tasks of
GPT-3.5-instruct).  Looking through the relative task samples, two typical
cases are depicted in \cref{fig:chat_vs_instruct}. It seems that the instruct
model is more prone to ignore the most recent instruction and try to achieve a
stale goal. The chat model is more adept at finding a target link through
scrolling and attempting (maybe a wrong link). However, sometimes, it requires
remarkable scrolling down steps to achieve the goal. In contrast, the instruct
model can make use of search again to quickly finish the new instruction.

\begin{table}[t]
	\centering
    \footnotesize
	\caption{Results of AgentLMs on WikiHow task set}
	\label{tab:agentlm_res}
	\begin{tabular}[b]{lcccccccc}
		\toprule[1.5pt]
		\multicolumn{1}{c}{\multirow{2}{*}{Model}} &	\multirow{2}{*}{Rwd} &	\multirow{2}{*}{SR} &	\multicolumn{2}{c}{Cross-Page} &	\multicolumn{2}{c}{In-Page} &	\multicolumn{2}{c}{QA} \\
		\cmidrule(lr){4-5}\cmidrule(lr){6-7}\cmidrule(lr){8-9}
                    &      &       & Rwd  & SR    & Rwd  & SR    & Rwd  & SR    \\
		\midrule
		AgentLM-7B  & 1.39 & 12.67 & 1.24 & 23.73 & 1.61 & 9.80  & 1.34 & 0.00  \\
		AgentLM-13B & 1.44 & 14.67 & 1.31 & 28.31 & 1.65 & 5.88  & 1.38 & 5.00  \\
		AgentLM-70B & 1.71 & 30.00 & 1.75 & 44.07 & 1.80 & 19.61 & 1.55 & 22.50 \\
		\bottomrule[1.5pt]
	\end{tabular}
\end{table}

\begin{figure}[t]
	\centering
	\begin{minipage}[b]{0.49\linewidth}
		\centering
		\captionof{table}{Ratios of invalid actions predicted by models.
		AgentLMs predict much less invalid actions than models from LLaMA 2.}
		\label{tab:inval_act_cmp}
		\begin{tabular}{cccc}
			\toprule[1.5pt]
			&	7B &	13B &	70B \\
			\midrule
			LLaMA-2 &	81.27 &	33.99 &	29.87 \\
			AgentLM &	15.44 &	13.45 &	3.12 \\
			\bottomrule[1.5pt]
		\end{tabular}
	\end{minipage}
	~
	\begin{minipage}[t]{0.49\linewidth}
		\centering
		\includegraphics[width=\linewidth]{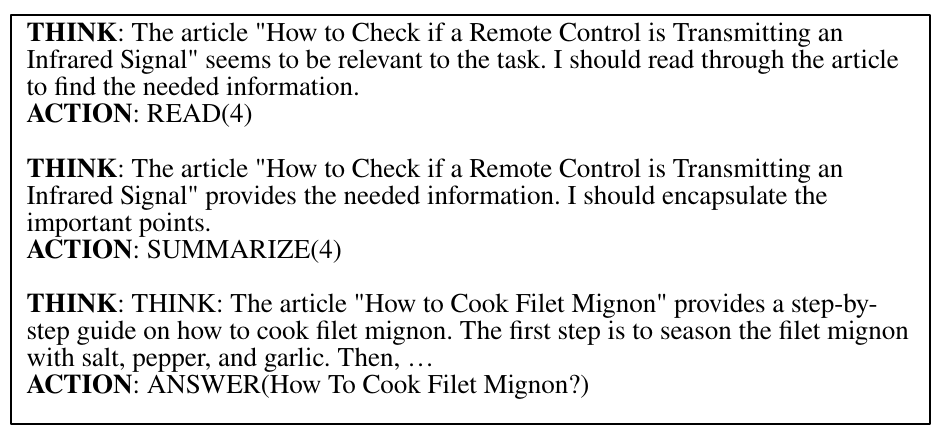}
		\caption{AgentLM-7B hallucinates invalid action types or returns the
		question itself as the final answer.}
		\label{fig:qa_fail_agtlm7b}
	\end{minipage}
\end{figure}

%

\paragraph{Performances of AgentLMs} The results of AgentLMs on the closed
world set are depicted in \cref{tab:agentlm_res}.  AgentLM models are further
fine-tuned based on LLaMA-2 chat models through AgentTuning to adapt to
decision-making tasks. They are expected to perform much better than the
original common-use models.  And the results do prove their effectiveness.
AgentLM models gain a remarkable improvement on WikiHow task set compared to
the original LLaMA-2 models, especially the 7B and 70B models. The strength of
AgentLMs mainly lie in the more capable ability to follow the output format,
\textit{i.e.}, less often outputting illegal actions like \texttt{SHARE} and
{\tt ANSWER: \ldots} (the correct format is \texttt{ANSWER(\ldots)}). This fact
is proven by the statistics in \cref{tab:inval_act_cmp}.  

\paragraph{Failure of AgentLM-7B on QA tasks} AgentLM-7B fails completely on QA
tasks. The trajectories (cases shown in \cref{fig:qa_fail_agtlm7b}) reveal that
AgentLM-7B often violates the output protocol and predicts invalid actions.
Sometimes, it also repeats the certain question as the answer, even a proper
summary has already been included in the thought. These phenomena reveal that
AgentLM-7B is still weak at format following.

\begin{figure}[t]
	\centering
	\includegraphics[width=\linewidth]{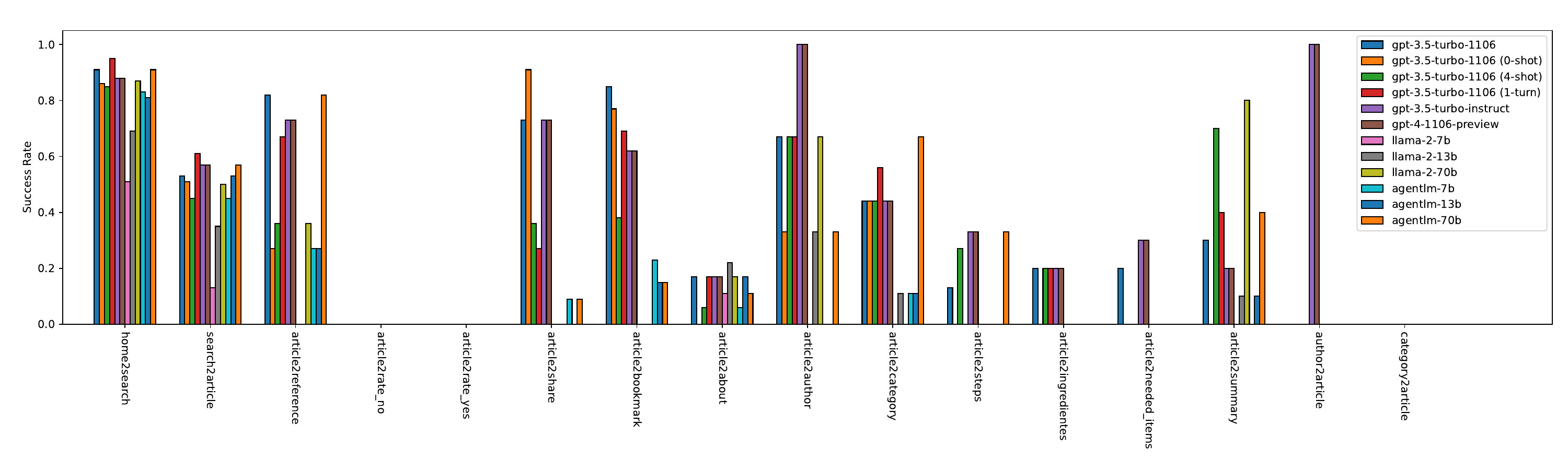}
	\caption{Success rates on single-stage sub-tasks}
	\label{fig:per_task_pfmc}
\end{figure}

\begin{wrapfigure}{R}{0.45\linewidth}
	\centering
	\includegraphics[width=\linewidth]{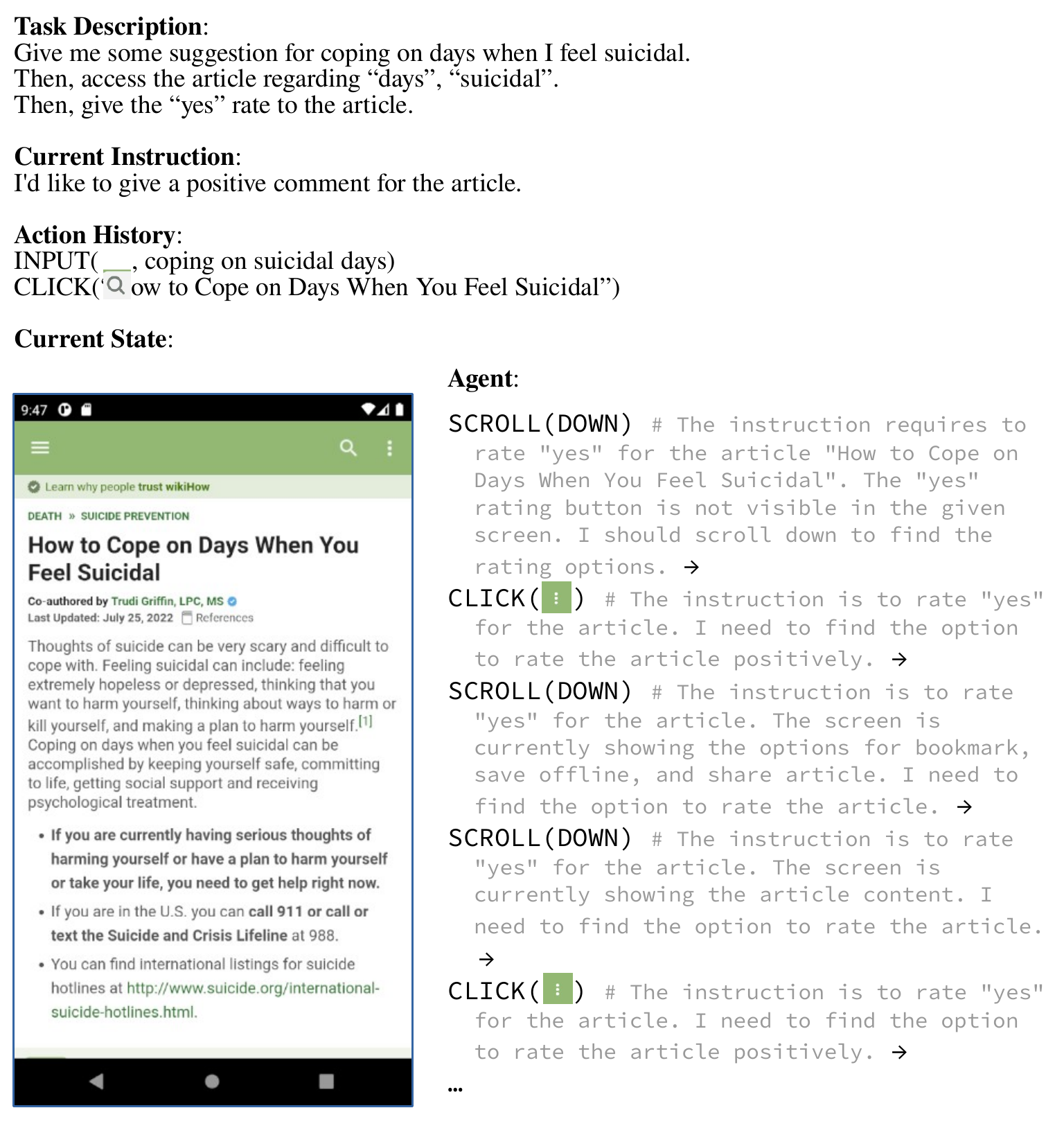}
	\caption{The agent has no belief to scroll down persistently.}
	\label{fig:gpt35_fail_rateyes}
\end{wrapfigure}

\paragraph{Results on single-stage sub-tasks} \cref{fig:per_task_pfmc}
visualizes the performance of the models on the single-stage sub-tasks
(introduced in \cref{subsec:t_gnrt_instrct_rwrt}). It is obvious that almost
all the models fail on sub-tasks \verb|article2rate_no|,
\verb|article2rate_yes|, \verb|author2article|, and \verb|category2article|.
This is caused by the need of long-term scrolling down to complete these tasks.
The buttons to give a rate for the article is at the page bottom. To find the
target article from the author information page or category content page also
requires scrolling down for a number of steps. However, the agents may lack the
belief to continuously scrolling down. They will try to click something aiming
at reaching the target quickly during scrolling, which results in an
inefficient policy and a failure after reaching the step number limit
(\textit{i.e.}, 15).  An example of GPT-3.5 is given in
\cref{fig:gpt35_fail_rateyes}. There are also shortcuts to complete these tasks
more efficiently, \textit{e.g.}, for \verb|article2rate_no| and
\verb|article2rate_yes|, the link to the reference list will bring the agent to
the very bottom of the article page near the rating buttons, and for
\verb|author2article| and \verb|category2article|, the agent can find the
target article through searching rather than seeking for it from the current
page. However, this may require some \textit{a priori} common-sense knowledge,
and few models can leverage this. If a learnable agent conducts some prior
explorations before testing (\textit{e.g.}, reinforcement learning training, or
building a grounded world model), it may leverage this \textit{a priori}
knowledge much better.  This point can be studied in future work.

\begin{figure}[htb]
	\centering
	\subfigure[VLM-based agent fails to ground the action
	accurately.
\label{fig:vlm_grd}]{\includegraphics[width=0.45\linewidth]{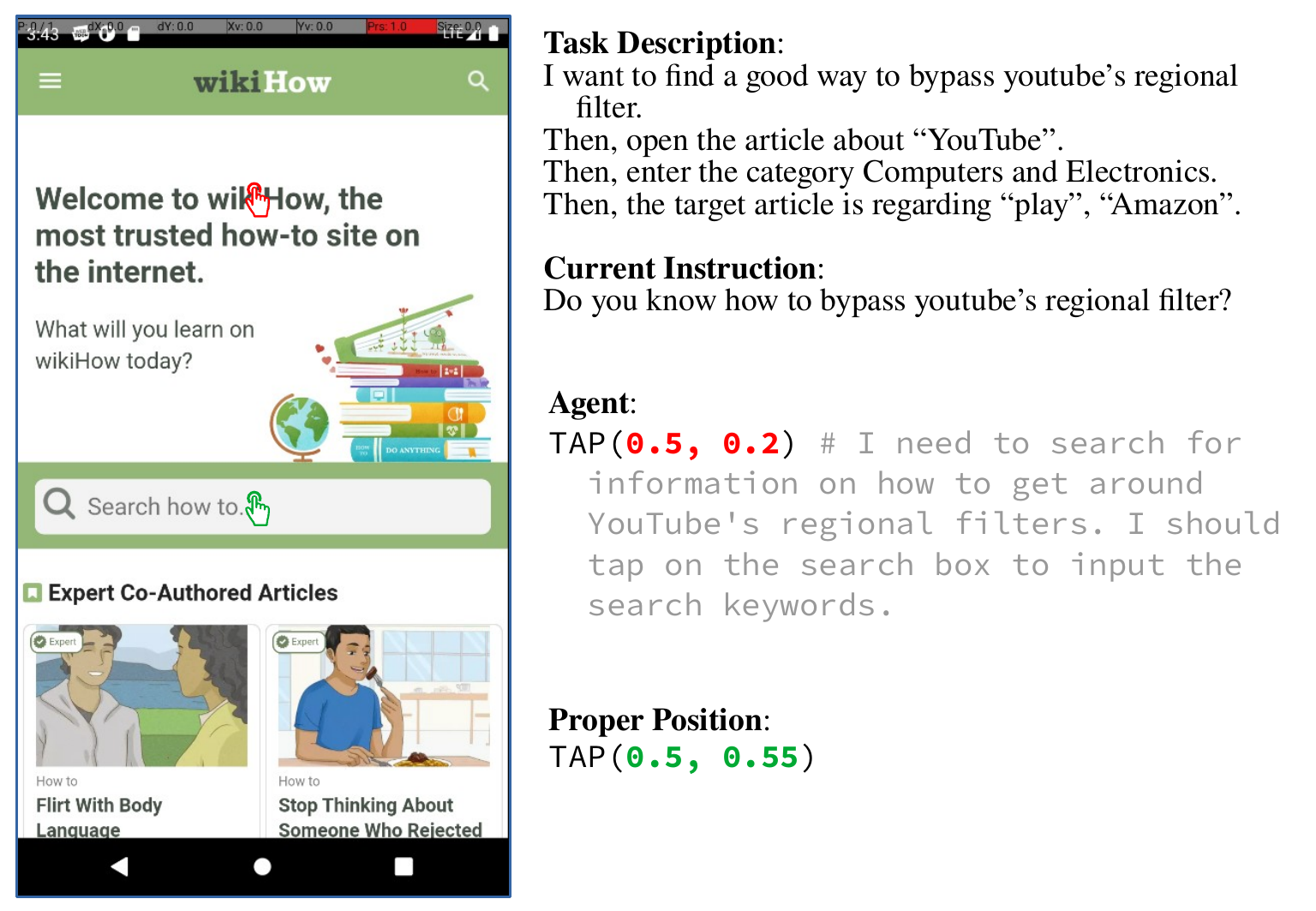}}
	~
	\subfigure[VLM-based agent is distracted by the
	popup.
\label{fig:vlm_dstrt}]{\includegraphics[width=0.45\linewidth]{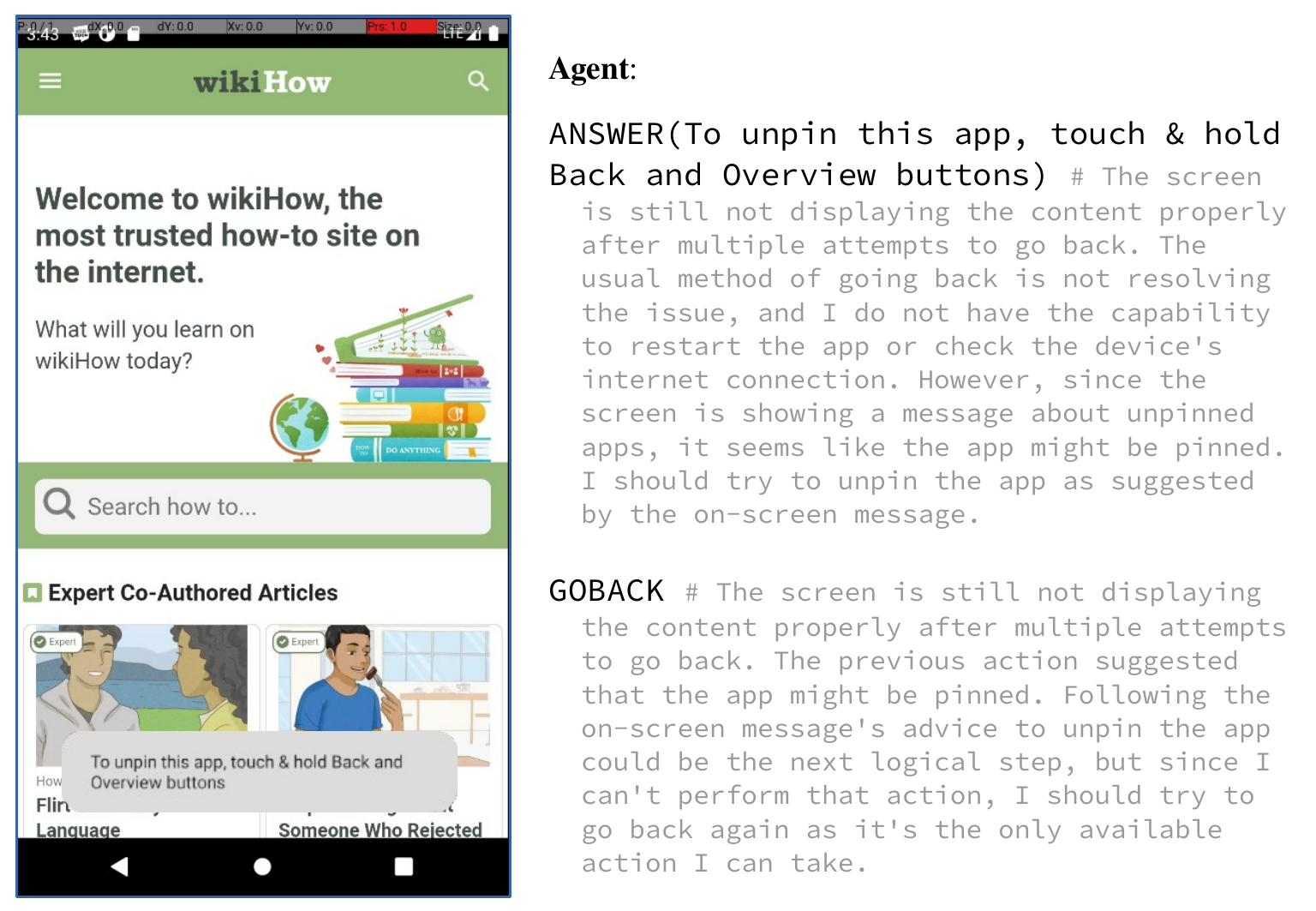}}
	\caption{Failure cases of VLMs}
	\label{fig:vlm_fail}
\end{figure}

\paragraph{Failure cases of VLMs} The VLM-based agent without SoM has a poor
capability to ground the action. As shown in \cref{fig:vlm_grd}, even though
the VLM predicts the plan correctly, it failed to accurately predicts the
coordinates. Meanwhile, VLM is more prone to be distracted by some notification
popups that are not present in the VH. For instance, in \cref{fig:vlm_dstrt},
the VLM is distracted by the popup about screen pinning and turns to try to
unpin the screen rather than focuses on completing the task goal.

%

\section{Limitations}
\label{sec:limit}

Despite the contribution of this work, there are still several limitations.
One of the concern is about data contamination of WikiHow task set. We should
claim that even though WikiHow is supposed to be a preferred corpus for LLM
pretraining, the involved data formats in WikiHow task set, regardless of
text-based or visual-based observations, are quite different from those that
will be used in pretraining. And the problem formulation also deviates from the
pretraining tasks of most common-use LLMs. However, there will be a latent risk
for QA tasks, \textit{i.e.}, the LLM may answer the question relying on its
parametric memory, rather than according to the underlying article.  Currently,
we haven't worked out an efficient solution to detect such contamination, and
we plan to leave it to future work for more efficient and effective
contamination detection method.  Besides, the performance of VLM-based agents
may also be affected by the SoM preprocessor. Fortunately, in accordance with
the inspection to the results of our adopted detectors, there are rare false
positives and essential missed true positives. Therefore, the performance of
the detector does not constitute a bottleneck of capability of the whole
SoM-based agent in our experiments.  In the end, more environments and task
sets are expected to be constructed to establish more challenging and
comprehensive benchmarks. It is also sincerely welcome that the community
contribute diverse environments and task sets based on Mobile-Env. Meanwhile,
We will continue maintaining and developing Mobile-Env platform as well. A
number of new features are going to be developed in the future.

\section{Impact statement}

This paper presents an interaction platform software Mobile-Env, along with the
open world and fixed world task sets, whose goal is to advance the field of GUI
interaction, especially LLM\slash VLM-driven GUI agents. With the developed software, new
\xxx{} GUI benchmarks can be established, and new agents can be built and
evaluated. There may be opportunity for vicious usage of the trained automatic
agents, however, no vicious actions can be performed directly through
Mobile-Env thanks to the isolated environments, to our best knowledge. 
No external data are involved in the open world set.
No crawler tools for building the fixed world set will be released. The 
fixed world WikiHow task set is collected purely from
publicly available data. To our best knowledge, there are no privacy data
included. There will be many other potential societal consequences of our work,
but none which we feel must be specifically highlighted here.

\section{Reviews from previous venue \& revision}

The previous reviewers argue that it is a common way to ``work with the
emulators thus building these environments'', consequently, we move the
implementation details from the main paper to the supplementary. Instead, we
elaborate on the justification for our design and the methodology we adopt to
handle the defects of existing benchmarks by our design. The previous reviewers
also expect tasks on more apps beyond WikiHow, thus, we crafted more tasks to
form an open world set and conducted experiments on it.

\section{Datasheet of fixed world task set}

\input{datasheet}

\end{document}

%% file: open-set-instructions.tex
\begin{longtable}{p{8.5cm}ccc}
	\caption{The instructions of the open world set.}
	\label{tab:open_wd_set_inst} \\

	\toprule[1.5pt]
	Instruction & Involved Apps & IR\textsuperscript{\textasteriskcentered} & II\textsuperscript{\dag} \\
	\midrule
	\endhead

	\bottomrule[1.5pt]
	\multicolumn{4}{l}{\footnotesize \textsuperscript{\textasteriskcentered}~Intermediate Rewards \quad \textsuperscript{\dag}~Intermediate Instructions} \\
	\endfoot

	I have multiple alarms to wake me up in the morning. Please check my alarms and tell me when I should get up finally tomorrow morning. & Alarm Clock & \ding{55} & \ding{55} \\
	Help me to check-in in Qidian\footnote{A reading app} app. & Qidian & \ding{55} & \ding{55} \\
	Help me to open the ``Do Not Disturb'' mode for my phone. & System Settings & \ding{55} & \ding{55} \\
	Turn off my alarm in the morning if it will rain tomorrow, otherwise leave it on. & Weather, Alarm Clock & \ding{55} & \ding{55} \\
	Check my last email from GitHub & Gmail, Notein\footnote{A noting app} & \ding{55} & \ding{51} \\
	Please disable all the notifications of Gmail for me. I want a free weekend. & System UI & \ding{55} & \ding{55} \\
	Send the illustration of Noelle I just downloaded to Lii Yihjiun through Gmail. & File Manager, Gmail & \ding{55} & \ding{51} \\
	Please help me to check the free spaces of my phone storage. & System Settings & \ding{55} & \ding{55} \\
	Turn on my alarm in the afternoon. & Alarm Clock & \ding{55} & \ding{55} \\
	Check my TOTP\footnote{Time-based One-Time Password} for GitHub and tell me the code. & Google Authenticator & \ding{55} & \ding{55} \\
	Turn off airplane mode for my phone. & System Settings & \ding{55} & \ding{55} \\
	Turn on airplane mode for my phone. & System Settings & \ding{55} & \ding{55} \\
	Tell me the Android version of my phone. & System Settings & \ding{55} & \ding{55} \\
	Install open-source 2FA\footnote{2-Factor Authentication} authenticator Aegis Authenticator from F-Droid\footnote{An open-source app store} for my phone. & F-Droid & \ding{55} & \ding{55} \\
	Install tool to connect my mobile with desktop KDE Connect from F-Droid for my phone. & F-Droid & \ding{55} & \ding{55} \\
	Install MPD client on Android M.A.L.P. from F-Droid for my phone. & F-Droid & \ding{55} & \ding{55} \\
	Install open-source bill management tool OpenMoneyBox from F-Droid for my phone. & F-Droid & \ding{55} & \ding{55} \\
	Install famous ``just works'' video and music player VLC from F-Droid for my phone. & F-Droid & \ding{55} & \ding{55} \\
	I have several funds and bills to be noted in OpenMoneyBox. Now open the app, enter the wizard, and help me to note the bills. & OpenMoneyBox & \ding{51} &\ding{51} \\
	Uninstall the app Google Authenticator and install open-source 2FA authenticator Aegis Authenticator from F-Droid instead for my phone. & System UI, F-Droid & \ding{51} &\ding{51} \\
	Uninstall the app VLC and install famous reimplementation of classical MPlayer MPV from F-Droid instead for my phone. & System UI, F-Droid & \ding{51} &\ding{51} \\
	Help me to set up a new alarm at 10:00 on Friday, Saturday, and Sunday. & Alarm Clock & \ding{55} & \ding{55} \\
	Help me to set up a new alarm at 11:59 on Thursday and Friday. & Alarm Clock & \ding{55} & \ding{55} \\
	Help me to set up a new alarm at 12:13 on Friday and Saturday. & Alarm Clock & \ding{55} & \ding{55} \\
	Help me to set up a new alarm at 13:00 on Monday and Tuesday. & Alarm Clock & \ding{55} & \ding{55} \\
	Help me to set up a new alarm at 14:34 on Thursday. & Alarm Clock & \ding{55} & \ding{55} \\
	Help me to set up a new alarm at 15:00 on Wednesday, Thursday, and Saturday. & Alarm Clock & \ding{55} & \ding{55} \\
	Help me to set up a new alarm at 16:00 on Tuesday. & Alarm Clock & \ding{55} & \ding{55} \\
	Help me to set up a new alarm at 7:00 on Monday and Tuesday. & Alarm Clock & \ding{55} & \ding{55} \\
	Help me to set up a new alarm at 8:00 on Friday. & Alarm Clock & \ding{55} & \ding{55} \\
	Help me to set up a new alarm at 9:00 on Monday. & Alarm Clock & \ding{55} & \ding{55} \\
	Uninstall the app KDE Connect on my phone. & System UI & \ding{55} & \ding{55} \\
	Uninstall the app M.A.L.P. on my phone. & System UI & \ding{55} & \ding{55} \\
	Uninstall the app OpenMoneyBox on my phone. & System UI & \ding{55} & \ding{55} \\
	Search an article to learn how to bake lobster tails. & WikiHow & \ding{51} & \ding{51} \\
	Search an article to learn how to deal with little sisters. & WikiHow & \ding{51} & \ding{51} \\
	Search an article to learn how to determine shipping costs. & WikiHow & \ding{51} & \ding{51} \\
	Search an article to learn how to file your nails. & WikiHow & \ding{51} & \ding{51} \\
	Search an article to learn how to hide short bangs. & WikiHow & \ding{51} & \ding{51} \\
	Search an article to learn how to how to behave yourself. & WikiHow & \ding{51} & \ding{51} \\
	Search an article to learn how to how to find true happiness and peace. & WikiHow & \ding{51} & \ding{51} \\
	Search an article to learn how to how to import from china into the usa. & WikiHow & \ding{51} & \ding{51} \\
	Search an article to learn how to how to make a homemade weight set. & WikiHow & \ding{51} & \ding{51} \\
	Search an article to learn how to make bottle cap art. & WikiHow & \ding{51} & \ding{51} \\
	Search an article to learn how to make pupusas. & WikiHow & \ding{51} & \ding{51} \\
	Search an article to learn how to rip dvd audio to mp3 using vlc media player. & WikiHow & \ding{51} & \ding{51} \\
	Search an article to learn how to wear high waisted shorts. & WikiHow & \ding{51} & \ding{51} \\
	Search an article to learn how to write a video game review. & WikiHow & \ding{51} & \ding{51} \\
	Search an article to learn how to introduce science to preschoolers. & WikiHow & \ding{51} & \ding{51} \\
	Search an article to learn how to make an electroscope. & WikiHow & \ding{51} & \ding{51} \\
	Search an article to learn how to play the palace card game. & WikiHow & \ding{51} & \ding{51} \\
	Search an article to learn how to rip jeggings. & WikiHow & \ding{51} & \ding{51} \\
	Search an article to learn how to transfer a vehicle tag in florida. & WikiHow & \ding{51} & \ding{51} \\
	Search an article to learn how to use scissors in gimp. & WikiHow & \ding{51} & \ding{51} \\
	Search the interstellar. & IMDB & \ding{51} & \ding{51} \\
	Navigate to the video part & IMDB & \ding{51} & \ding{51} \\
	Search the first two topics about Hongloumeng in Douban & Douban & \ding{51} & \ding{51} \\
	Get in the douban group of the book named Sanguoyanyi & Douban & \ding{51} & \ding{51} \\
	Return the douban score of book named Sanguoyanyi & Douban & \ding{51} & \ding{51} \\
	Tell me the 4th movie in the Douban Top250 movies list. & Douban & \ding{51} & \ding{51} \\
	Search the Breaking Bad TV show on IMDb & IMDB & \ding{51} & \ding{51} \\
	Search the 2002 verison spiderman movie and check the rate for it. & IMDB & \ding{51} & \ding{51} \\
	Search the book named Sanguoyanyi. & Douban & \ding{51} & \ding{51} \\
	Return the director of the TV named Zhenhuanzhuan & Douban & \ding{51} & \ding{51} \\
	Open the address table. & Ele.me & \ding{51} & \ding{51} \\
	Help me order takeout. & Ele.me & \ding{51} & \ding{51} \\
	Help me order takeout. & Ele.me & \ding{51} & \ding{51} \\
	Select one of my previous orders from my entire order history and place a new order of the same items to help me order takeout again. & Ele.me & \ding{51} & \ding{51} \\
	Help me order a cup of my favorite milk tea by searching for it. & Ele.me & \ding{51} & \ding{51} \\
	Help me order a cup of my favorite milk tea by searching for Coco (the store name). & Ele.me & \ding{51} & \ding{51} \\
	Help me select takeout that can be delivered within 30 minutes using the filters on the homepage and place the order. & Ele.me & \ding{51} & \ding{51} \\
	View my order history. & Ele.me & \ding{51} & \ding{51} \\
	View my shopping cart. & Ele.me & \ding{51} & \ding{51} \\
	View the messages page. & Ele.me & \ding{51} & \ding{51} \\
\end{longtable}

%% file: chat_prompt.tex
\begin{longtable}{rp{11.5cm}}
	\caption{Prompt for text LLMs}
	\label{tab:text_llm_prmpt} \\

	\toprule[1.5pt]
	\endhead

	\bottomrule[1.5pt]
	\endfoot

	\textcolor{violet}{System: }
	& Given several information about the task and the screen at the current step, I need to take an appropriate action according to the given information to finish the task. Available actions are: \\
	&  \\
	& {\tt INPUT(element\_id, text) \# You can input something into search area through this action} \\
	& {\tt CLICK(element\_id) \# You can click on some clickable element through this action} \\
	& {\tt LONG\_CLICK(element\_id) \# You can long lick on some clickable element through this action} \\
	& {\tt SCROLL(direction) \# You can scroll up\slash down to browse the long page through this action} \\
	& {\tt ANSWER(text) \# You can generate an answer to me through this action} \\
	& {\tt GOBACK \# You can return to the last page through this action} \\
	&  \\
	& At every time decision, you need first output a line of think, and then output a line of action. For example: \\
	&  \\
	& {\tt THINK:} I need to have a thinking before I take my action. \\
	& {\tt ACTION: ANSWER(I can take any available action, e.g., give an answer.)} \\
	&  \\
	& The above is just an example and takes no any relation with the following inputs. Now, take your mission. \\
	\textcolor{caoblue}{User: }
	& {\tt {`}{`}{`}} \\
	& {\tt Task:} \\
	& {\tt Look up how to conceal gauges.} \\
	& {\tt Then, access the article regarding gauges.} \\
	& {\tt Screen:} \\
	& {\tt <button alt="Open navigation drawer" id="0" clickable="true"></button>} \\
	& {\tt <img class="wikihow toolbar logo" id="1" clickable="false">} \\
	& {\tt <img class="search button" alt="Search" id="2" clickable="true">} \\
	& {\tt <div class="webView" id="3" clickable="true"></div>} \\
	& {\tt <div class="statusBarBackground" id="4" clickable="false"></div>} \\
	& {\tt Instruction:} \\
	& {\tt } \\
	& {\tt Action History:} \\
	& {\tt } \\
	& {\tt {-}{-}{-}} \\
	\textcolor{dullred}{Assistant: }
	& {\tt THINK:} I should search "hide gauges" to know how to conceal gauges. \\
	& {\tt ACTION: INPUT(2, hide gauges)} \\
	\textcolor{caoblue}{User: }
	& {\tt {`}{`}{`}} \\
	& {\tt Task:} \\
	& {\tt How to style ruby rose hair?} \\
	& {\tt Then, open the article regarding ruby.} \\
	& {\tt Then, check the about page for the reliability of WikiHow.} \\
	& {\tt Screen:} \\
	& {\tt <button alt="Open navigation drawer" id="0" clickable="true"></button>} \\
	& {\tt <input class="search src text" value="Do ruby rose hair " type="text" id="1" clickable="true">} \\
	& {\tt <img class="search close btn" alt="Clear query" id="2" clickable="true">} \\
	& {\tt <div id="3" clickable="true">How to Do Ruby Rose Hair</div>} \\
	& {\tt <div id="4" clickable="true">• </div>} \\
	& {\tt <p id="5" clickable="true">41,446 views</p>} \\
	& {\tt <div id="6" clickable="true">• </div>} \\
	& {\tt <p id="7" clickable="true">Updated</p>} \\
	& {\tt <p id="8" clickable="true">2 years ago</p>} \\
	& {\tt <div id="9" clickable="true">• </div>} \\
	& {\tt <p id="10" clickable="true">Expert Co-Authored</p>} \\
	& {\tt <div id="11" clickable="true">How to Dye Your Hair Rose Gold</div>} \\
	& {\tt <div id="12" clickable="true">• </div>} \\
	& {\tt <p id="13" clickable="true">48,548 views</p>} \\
	& {\tt <div id="14" clickable="true">• </div>} \\
	& {\tt <p id="15" clickable="true">Updated</p>} \\
	& {\tt <p id="16" clickable="true">3 years ago</p>} \\
	& {\tt <div id="17" clickable="true">• </div>} \\
	& {\tt <p id="18" clickable="true">Expert Co-Authored</p>} \\
	& {\tt <div id="19" clickable="true">How to Do a Hair Mask for Frizzy Hair</div>} \\
	& {\tt <div id="20" clickable="true">• </div>} \\
	& {\tt <p id="21" clickable="true">29,415 views</p>} \\
	& {\tt <div id="22" clickable="true">• </div>} \\
	& {\tt <p id="23" clickable="true">Updated</p>} \\
	& {\tt <p id="24" clickable="true">2 years ago</p>} \\
	& {\tt <div id="25" clickable="true">• </div>} \\
	& {\tt <p id="26" clickable="true">Expert Co-Authored</p>} \\
	& {\tt <div id="27" clickable="true">How to Do 1920s Hair</div>} \\
	& {\tt <div id="28" clickable="true">• </div>} \\
	& {\tt <p id="29" clickable="true">194,834 views</p>} \\
	& {\tt <div id="30" clickable="true">• </div>} \\
	& {\tt <p id="31" clickable="true">Updated</p>} \\
	& {\tt <p id="32" clickable="true">1 year ago</p>} \\
	& {\tt <div id="33" clickable="true">• </div>} \\
	& {\tt <p id="34" clickable="true">Expert Co-Authored</p>} \\
	& {\tt <div id="35" clickable="true">How to Do 50s Hair</div>} \\
	& {\tt <div id="36" clickable="true">• </div>} \\
	& {\tt <p id="37" clickable="true">39,890 views</p>} \\
	& {\tt <div id="38" clickable="true">• </div>} \\
	& {\tt <p id="39" clickable="true">Updated</p>} \\
	& {\tt <p id="40" clickable="true">1 year ago</p>} \\
	& {\tt <div id="41" clickable="true">• </div>} \\
	& {\tt <p id="42" clickable="true">Expert Co-Authored</p>} \\
	& {\tt <div id="43" clickable="true">How to Do 70s Hair</div>} \\
	& {\tt <div id="44" clickable="true">• </div>} \\
	& {\tt <p id="45" clickable="true">69,458 views</p>} \\
	& {\tt <div id="46" clickable="true">• </div>} \\
	& {\tt <p id="47" clickable="true">Updated</p>} \\
	& {\tt <p id="48" clickable="true">1 year ago</p>} \\
	& {\tt <div id="49" clickable="true">• </div>} \\
	& {\tt <p id="50" clickable="true">Expert Co-Authored</p>} \\
	& {\tt <div id="51" clickable="true">How to Do Goku Hair</div>} \\
	& {\tt <div id="52" clickable="true">• </div>} \\
	& {\tt <p id="53" clickable="true">44,049 views</p>} \\
	& {\tt <div class="statusBarBackground" id="54" clickable="false"></div>} \\
	& {\tt Instruction:} \\
	& {\tt "How to Do Ruby Rose Hair" may be helpful for us.} \\
	& {\tt Action History:} \\
	& {\tt INPUT(2, do ruby rose hair) \# I should search "do ruby rose hair" to know how to style ruby rose hair.} \\
	& {\tt {-}{-}{-}} \\
	\textcolor{dullred}{Assistant: }
	& {\tt THINK:} The instruction requires to read the article "How to Do Ruby Rose Hair". Element 3 is "How to Do Ruby Rose Hair" and is clickable. I should click it to access the article. \\
	& {\tt ACTION: CLICK(3)} \\
\end{longtable}

%% file: mm_prompt.tex
\begin{longtable}{rp{11.5cm}}
	\caption{Prompt for VLMs}
	\label{tab:vis_llm_prmpt} \\

	\toprule[1.5pt]
	\endhead

	\bottomrule[1.5pt]
	\endfoot

	\textcolor{violet}{System: }
	& Given several information about the task and the screen at the current step, I need to take an appropriate action according to the given information to finish the task. Available actions are: \\
	&  \\
	& {\tt TAP(element\_id) \# You can touch a marked element on the screen by its number id.} \\
	& {\tt LONG\_TAP(element\_id) \# You can long touch a marked element on the screen by its number id.} \\
	& {\tt SCROLL(direction) \# You can scroll up\slash down to browse the long page through this action} \\
	& {\tt TYPE(text) \# You can type in something through keyboard.} \\
	& {\tt ANSWER(text) \# You can generate an answer to me through this action.} \\
	& {\tt GOBACK \# You can return to the last page through this action.} \\
	&  \\
	& At every time decision, you need first output a line of think, and then output a line of action. For example: \\
	&  \\
	& {\tt THINK:} I need to have a thinking before I take my action. \\
	& {\tt ACTION: ANSWER(I can take any available action, e.g., give an answer.)} \\
	&  \\
	& The above is just an example and takes no any relation with the following inputs. Now, take your mission. \\
	\textcolor{caoblue}{User: }
	& {\tt Task:} \\
	& {\tt Look up how to conceal gauges.} \\
	& {\tt Then, access the article regarding gauges.} \\
	& {\tt Instruction:} \\
	& {\tt } \\
	& {\tt Action History:} \\
	& {\tt TAP(13) \# I should search "hide gauges" to know how to conceal gauges. I can tap on the search box to input the search keywords.} \\
	& {\tt Screen:} \\
	& {\tt \$\{imagef:prompts/mm\_prompt/eg1\_som.png:280:500\}} \\
	\textcolor{dullred}{Assistant: }
	& {\tt THINK:} I should search "hide gauges" to know how to conceal gauges. I have selected the search box and need to type the search keywords. \\
	& {\tt ACTION: TYPE(hige gauges)} \\
	\textcolor{caoblue}{User: }
	& {\tt Task:} \\
	& {\tt How to style ruby rose hair?} \\
	& {\tt Then, open the article regarding ruby.} \\
	& {\tt Then, check the about page for the reliability of WikiHow.} \\
	& {\tt Instruction:} \\
	& {\tt "How to Do Ruby Rose Hair" may be helpful for us.} \\
	& {\tt Action History:} \\
	& {\tt TAP(13) \# I should search "Do ruby rose hair" to know how to style ruby rose hair. I can tap on the search box to input the search keywords.} \\
	& {\tt TYPE(Do ruby rose hair) \# I should search "Do ruby rose hair" to know how to style ruby rose hair. I have selected the search box and need to type the search keywords.} \\
	& {\tt Screen:} \\
	& {\tt \$\{imagef:prompts/mm\_prompt/eg2\_som.png:280:500\}} \\
	\textcolor{dullred}{Assistant: }
	& {\tt THINK:} The instruction requires to read the article "How to Do Ruby Rose Hair". Thus, I should click on this link. \\
	& {\tt ACTION: TAP(3)} \\
\end{longtable}

%% file: datasheet.tex
\subsection{Motivation}

\begin{itemize}
	\item {\bf For what purpose was the dataset created?} Was there a specific
		task in mind? Was there a specific gap that needed to be filled? Please
		provide a description.

		Mobile-Env platform is designed to assist in building a \xxx{} GUI
		interaction benchmark that holds reliable evaluation, isolated \&
		controllable environments, intermediate rewards, and intermediate
		instructions. WikiHow task set is created based on Mobile-Env to
		provide the first truly reproducible and comparable GUI benchmark.

	\item {\bf Who created the dataset (\textit{e.g.}, which team, research
		group) and on behalf of which entity (\textit{e.g.}, company,
	institution, organization)?}
	
		Danyang Zhang, Zhennan Shen, Rui Xie, Situo Zhang, Tianbao Xie, Zihan
		Zhao, Siyuan Chen, Lu Chen, Hongshen Xu, Ruisheng Cao, and Kai Yu
		create Mobile-Env platform and WikiHow task set. The authors are mainly
		from X-Lance Lab, SJTU.

%

\end{itemize}

\subsection{Composition}

\begin{itemize}
	\item {\bf What do the instances that comprise the dataset represent
		(\textit{e.g.}, documents, photos, people, countries)?} Are there
		multiple types of instances (\textit{e.g.}, movies, users, and ratings;
		people and interactions between them; nodes and edges)? Please provide
		a description.

		The task definitions of WikiHow task set are provided in the text
		format of Protocol Buffers\footnote{\url{https://protobuf.dev/}}. They
		can be directly loaded and parsed by Mobile-Env. The app data of
		WikiHow are given as the captured HTTP flows. All types of media
		(\textit{e.g.}, web pages, images, and videos) are provided as raw
		bytes for convenience during replay.

	\item {\bf How many instances are there in total (of each type, if
		appropriate)?}

		We totally release 150 task definitions for WikiHow task set and report
		results on it.  There are totally 856,045 resources crawled from
		WikiHow website, in which there are 107,448 distinct web pages.
		
	\item {\bf Does the dataset contain all possible instances or is it a
		sample (not necessarily random) of instances from a larger set?} If the
		dataset is a sample, then what is the larger set? Is the sample
		representative of the larger set (\textit{e.g.}, geographic coverage)?
		If so, please describe how this representativeness was validated\slash
		verified. If it is not representative of the larger set, please
		describe why not (\textit{e.g.}, to cover a more diverse range of
		instances, because instances were withheld or unavailable).

		There are over 340,000 articles on WikiHow, from which about 100,000
		ones are crawled. The pages are accessed in breadth-first order to
		mimic the browsing behavior of human users and agents. We argue that
		the current saved flow data are sufficient for the exploration of
		agents in the app. The task definitions are instantiated with the
		sampled keyword information from the crawled WikiHow resources. 

	\item {\bf What data does each instance consist of?} ``Raw'' data
		(\textit{e.g.}, unprocessed text or images) or features? In either case,
		please provide a description.

		The app data of WikiHow are stored as raw flow. Each flow file consists
		of the HTTP headers in text and the payload in raw bytes.

	\item {\bf Is there a label or target associated with each instance?} If so,
		please provide a description.

		The task configuration files in WikiHow task set define task goals,
		intermediate instructions, and rewards.

	\item {\bf Is any information missing from individual instances?} If so,
		please provide a description, explaining why this information is
		missing (\textit{e.g.}, because it was unavailable). This does not
		include intentionally removed information, but might include,
		\textit{e.g.}, redacted text.

		No.

	\item {\bf Are there any errors, sources of noise, or redundancies in the
		dataset?} If so, please provide a description.

		Redundancies may exist in the crawled app data of WikiHow task set, as
		some URLs may point to the same resource.

	\item {\bf Is the dataset self-contained, or does it link to or otherwise
rely on external resources (\textit{e.g.}, websites, tweets, other datasets)?}
If it links to or relies on external resources, a) are there guarantees that
they will exist, and remain constant, over time; b) are there official archival
versions of the complete dataset (\textit{i.e.}, including the external
resources as they existed at the time the dataset was created); c) are there
any restrictions (\textit{e.g.}, licenses, fees) associated with any of the
external resources that might apply to a dataset consumer? Please provide
descriptions of all external resources and any restrictions associated with
them, as well as links or other access points, as appropriate.

		WikiHow task set is considered self-contained. All the required app
		data for WikiHow task set are saved.

	\item {\bf Does the dataset contain data that might be considered
		confidential (\textit{e.g.}, data that is protected by legal privilege
	or by doctor–patient confidentiality, data that includes the content of
	individuals' non-public communications)?} If so, please provide a
	description.

		No. All the crawled app data are published publicly on WikiHow website.

	\item {\bf Does the dataset contain data that, if viewed directly, might be
		offensive, insulting, threatening, or might otherwise cause anxiety?}
		If so, please describe why.

		No. All the crawled app data have already been published according to
		the content policy of WikiHow website.

	\item {\bf Does the dataset identify any subpopulations (\textit{e.g.}, by
		age, gender)?} If so, please describe how these subpopulations are
		identified and provide a description of their respective distributions
		within the dataset.

		No.

	\item {\bf Is it possible to identify individuals (\textit{i.e.}, one or
		more natural persons), either directly or indirectly (\textit{i.e.}, in
	combination with other data) from the dataset?} If so, please describe how.

		The article authors may be identified through the pages of article and
		author information, which are publicly available on WikiHow website.

	\item {\bf Does the dataset contain data that might be considered sensitive
			in any way (\textit{e.g.}, data that reveals race or ethnic
				origins, sexual orientations, religious beliefs, political
				opinions or union memberships, or locations; financial or
				health data; biometric or genetic data; forms of government
		identification, such as social security numbers; criminal history)?} If
		so, please provide a description.

		No. All the crawled app data have already been published according to
		the content policy of WikiHow website.

\end{itemize}

\subsection{Collection Process}

\begin{itemize}
	\item {\bf How was the data associated with each instance acquired?} Was
		the data directly observable (\textit{e.g.}, raw text, movie ratings),
		reported by subjects (\textit{e.g.}, survey responses), or indirectly
		inferred\slash derived from other data (\textit{e.g.}, part-of-speech
		tags, model-based guesses for age or language)?  If the data was
		reported by subjects or indirectly inferred\slash derived from other
		data, was the data validated\slash verified? If so, please describe
		how.

		All the crawled app data and information used in the task definitions
		of WikiHow task set are directly observable on WikiHow website.

	\item {\bf What mechanisms or procedures were used to collect the data
		(\textit{e.g.}, hardware apparatuses or sensors, manual human curation,
	software programs, software APIs)?} How were these mechanisms or procedures
	validated?
	
		Mobile-Env platform is developed with Python based on Android Emulator.
		Several auxiliary tools and scripts are created with Tcl, JavaScript,
		\textit{etc}. The task definitions of WikiHow task set are generated
		with the template toolkit (introduced in \cref{sub:misc_tool}) of
		Mobile-Env. The app data of WikiHow task set are crawled with a Python
		crawler on a host with 8 CPU threads and 15 GiB memory.

	\item {\bf If the dataset is a sample from a larger set, what was the
		sampling strategy (\textit{e.g.}, deterministic, probabilistic with
	specific sampling probabilities)?}

		The task definitions of WikiHow task set are combined from the
		single-stage task definitions randomly under the constraint that
		guarantees the target page in the succeeding stage is directly
		referenced on the page in the preceding stage. The app data are crawled
		in breadth-first order starting from the home page to mimic the
		browsing behavior of agents and offer the largest coverage of the
		agents' exploration.

	\item {\bf Who was involved in the data collection process (\textit{e.g.},
		students, crowdworkers, contractors) and how were they compensated
	(\textit{e.g.}, how much were crowdworkers paid)?}

		All the development of platform, app data collection, and task
		configuration creation are completed by the authors.

	\item {\bf Over what timeframe was the data collected?} Does this timeframe
		match the creation timeframe of the data associated with the instances
		(\textit{e.g.}, recent crawl of old news articles)? If not, please
		describe the timeframe in which the data associated with the instances
		was created.

		The app data of WikiHow task set are collected from WikiHow website
		from Jul.\ 6th, 2022 to Aug.\ 18th, 2022. Wiki articles published on
		the website before the date may be included.


	\item {\bf Did you collect the data from the individuals in question
		directly, or obtain it via third parties or other sources (e.g.,
	websites)?}

		The app data including the published wiki articles and public author
		information are crawled from WikiHow website.

	\item {\bf Were the individuals in question notified about the data
		collection?} If so, please describe (or show with screenshots or other
		information) how notice was provided, and provide a link or other
		access point to, or otherwise reproduce, the exact language of the
		notification itself.

		The individuals as the wiki article authors are not notified.

	\item {\bf Did the individuals in question consent to the collection and
		use of their data?} If so, please describe (or show with screenshots or
		other information) how consent was requested and provided, and provide
		a link or other access point to, or otherwise reproduce, the exact
		language to which the individuals consented.

		Only the publicly available data on WikiHow website are crawled. We
		obey the terms of use and \texttt{robots.txt} of WikiHow website. Thus,
		it is considered that the individuals as the wiki article authors
		consent to the collection and usage of their works.

	\item {\bf If consent was obtained, were the consenting individuals
		provided with a mechanism to revoke their consent in the future or for
	certain uses?} If so, please provide a description, as well as a link or
	other access point to the mechanism (if appropriate).

		The crawled app data are regarded as a snapshot of WikiHow website.
		Currently, no mechanism for revocation is designed. However, if there
		is really a need to revoke the particular resources, the individual can
		contact the paper authors directly.

\end{itemize}

\subsection{Uses}

\begin{itemize}
	\item {\bf Has the dataset been used for any tasks already?} If so, please
		provide a description.

		Agents based on different LLMs have been evaluated on WikiHow task set
		in this paper.

	\item {\bf Is there a repository that links to any or all papers or systems
		that use the dataset?} If so, please provide a link or other access
		point.
		
		Currently, no. We plan to develop a website tracking and indexing the
		available task sets for Mobile-Env.

	\item {\bf Is there anything about the composition of the dataset or the
		way it was collected and preprocessed\slash cleaned\slash labeled that
	might impact future uses?} For example, is there anything that a dataset
	consumer might need to know to avoid uses that could result in unfair
	treatment of individuals or groups (\textit{e.g.}, stereotyping, quality of
	service issues) or other risks or harms (\textit{e.g.}, legal risks,
	financial harms)? If so, please provide a description. Is there anything a
	dataset consumer could do to mitigate these risks or harms?

		No.

\end{itemize}

\subsection{Distribution}

\begin{itemize}
	\item {\bf Will the dataset be distributed to third parties outside of the
		entity (\textit{e.g.}, company, institution, organization) on behalf of
	which the dataset was created?} If so, please provide a description.

		Yes. Both Mobile-Env platform and WikiHow task set are open-sourced.

	\item {\bf How will the dataset will be distributed (\textit{e.g.}, tarball
		on website, API, GitHub)?} Does the dataset have a digital object
		identifier (DOI)?

		The platform is open-sourced at GitHub. The task set is released at
		Hugging Face. We do not apply for a DOI.

	\item {\bf When will the dataset be distributed?}

		Both platform and task set have already been made public.

	\item {\bf Will the dataset be distributed under a copyright or other
		intellectual property (IP) license, and\slash or under applicable terms
	of use (ToU)?} If so, please describe this license and\slash or ToU, and
	provide a link or other access point to, or otherwise reproduce, any
	relevant licensing terms or ToU, as well as any fees associated with these
	restrictions.

		Both Mobile-Env platform and WikiHow task set are open-sourced under
		Apached-2.0 license.

	\item {\bf Have any third parties imposed IP-based or other restrictions on
		the data associated with the instances?} If so, please describe these
		restrictions, and provide a link or other access point to, or otherwise
		reproduce, any relevant licensing terms, as well as any fees associated
		with these restrictions.

		The copyright of the app data in WikiHow task set is owned by the
		original wiki authors and WikiHow website. No other third parties own
		the copyright of Mobile-Env platform and the task definitions of
		WikiHow task set.

	\item {\bf Do any export controls or other regulatory restrictions apply to
		the dataset or to individual instances?} If so, please describe these
		restrictions, and provide a link or other access point to, or otherwise
		reproduce, any supporting documentation.

		No.

\end{itemize}

\subsection{Maintenance}

\begin{itemize}
	\item {\bf Who will be supporting\slash hosting\slash maintaining the
		dataset?}

		The authors will support, host, and maintain both Mobile-Env platform
		and WikiHow task set.

	\item {\bf How can the owner\slash curator\slash manager of the dataset be
		contacted (\textit{e.g.}, email address)?}

		Issues and discussions on GitHub and Hugging Face are welcome. One can
		also seek help from Danyang Zhang ({\tt zhang-dy20@sjtu.edu.cn}), Lu
		Chen ({\tt chenlusz@sjtu.edu.cn}), and Kai Yu ({\tt
		kai.yu@sjtu.edu.cn}).

	\item {\bf Is there an erratum?} If so, please provide a link or other
		access point.

		Currently, no. Errata will be announced if there are any in the future.

	\item {\bf Will the dataset be updated (\textit{e.g.}, to correct labeling
		errors, add new instances, delete instances)?} If so, please describe
		how often, by whom, and how updates will be communicated to dataset
		consumers (\textit{e.g.}, mailing list, GitHub)?

		Mobile-Env platform will be continuously developed and maintained.
		Updates will be released on GitHub from time to time. Currently, there
		is no remarkable updating plan for WikiHow task set. Errata may be
		released to correct errors if there are any in the future.

	\item {\bf If the dataset relates to people, are there applicable limits on
		the retention of the data associated with the instances (\textit{e.g.},
	were the individuals in question told that their data would be retained for
	a fixed period of time and then deleted)?} If so, please describe these
	limits and explain how they will be enforced.

		No.

	\item {\bf Will older versions of the dataset continue to be
		supported\slash hosted\slash maintained?} If so, please describe how.
		If not, please describe how its obsolescence will be communicated to
		dataset consumers.

		Old versions of Mobile-Env and WikiHow task set can be accessed through
		GitHub and Hugging Face, respectively.

	\item {\bf If others want to extend\slash augment\slash build on\slash
		contribute to the dataset, is there a mechanism for them to do so?} If
		so, please provide a description. Will these contributions be
		validated/verified? If so, please describe how. If not, why not? Is
		there a process for communicating\slash distributing these
		contributions to dataset consumers? If so, please provide a
		description.

		We sincerely welcome that one can contribute features or report bugs
		for Mobile-Env through the mechanisms like pull request, issues,
		\textit{etc}.\ on GitHub. If new environments or task sets are crafted,
		it will be welcome that the creators notify the authors through e-mail
		or GitHub, so that we can update the indices of available task sets.
		
\end{itemize}